# Interactive Cost Configuration Over Decision Diagrams

**Henrik Reif Andersen**                                          HRA@CONFIGIT.COM
*Configit A/S*
*DK-2100 Copenhagen, Denmark*

**Tarik Hadzic**                                                  T.HADZIC@4C.UCC.IE
*Cork Constraint Computation Centre*
*University College Cork*
*Cork, Ireland*

**David Pisinger**                                               PISINGER@MAN.DTU.DK
*DTU Management*
*Technical University of Denmark*
*DK-2800 Kgs. Lyngby, Denmark*

## Abstract

In many AI domains such as product configuration, a user should interactively specify a solution that must satisfy a set of constraints. In such scenarios, offline compilation of feasible solutions into a tractable representation is an important approach to delivering efficient backtrack-free user interaction online. In particular, *binary decision diagrams* (BDDs) have been successfully used as a compilation target for product and service configuration. In this paper we discuss how to extend BDD-based configuration to scenarios involving *cost functions* which express user preferences.

We first show that an efficient, robust and easy to implement extension is possible if the cost function is *additive*, and feasible solutions are represented using *multi-valued decision diagrams* (MDDs). We also discuss the effect on MDD size if the cost function is non-additive or if it is encoded explicitly into MDD. We then discuss interactive configuration in the presence of multiple cost functions. We prove that even in its simplest form, multiple-cost configuration is NP-hard in the input MDD. However, for solving two-cost configuration we develop a pseudo-polynomial scheme and a fully polynomial approximation scheme. The applicability of our approach is demonstrated through experiments over real-world configuration models and product-catalogue datasets. Response times are generally within a fraction of a second even for very large instances.

## 1. Introduction

Interactively specifying a solution that must satisfy a number of combinatorial restrictions is an important problem in many AI domains related to decision making: from buying a product online, selling an insurance policy to setting up a piece of equipment. Solutions are often modeled as assignments to variables over which constraints are imposed. When assigning variables without sufficient guidance, a user might be forced to backtrack, since some of the choices he made cannot be extended in a way that would satisfy all of the succeeding constraints. To improve the usability of interaction it is therefore important to indicate to a user all values that participate in at least one remaining solution. If a user is assigning only such values he is guaranteed to be able to reach any feasible solution while never being forced to backtrack. We refer to the task of computing such values as *calculating valid domains* (CVD). Since this is a computationally challenging (NP-hard) problem, and short execution times are important in an interactive setting, it has been suggested to *compile* offline (prior to user interaction) the set of all feasible solutions into a representation form that supports efficient execution of CVD during online interaction.

Møller, Andersen, and Hulgaard (2002) and Hadzic, Subbarayan, Jensen, Andersen, Møller, and Hulgaard (2004) investigated such an approach by using *binary decision diagrams* (BDDs) as a compilation target. BDDs are one of the data-structures investigated in the *knowledge compilation* community which preprocess original problem formulations into more tractable representations to enhance solving the subsequent tasks. CVD is just one of such tasks occurring in the configuration domain. Knowledge compilation has been successfully applied to a number of other areas such as planning, diagnosis, model checking etc. Beside BDDs, a number of other structures, such as various sublanguages of negation normal forms (NNFs) (Darwiche & Marquis, 2002), AND/OR diagrams (Mateescu, Dechter, & Marinescu, 2008), finite state automata (Vempaty, 1992; Amilhastre, Fargier, & Marquis, 2002) and various extensions of decision diagrams (Drechsler, 2001; Wegener, 2000; Meinel & Theobald, 1998) are used as compilation targets. Some of them are suitable for interactive configuration as well. In particular, Vempaty (1992) suggested compiling constraints into an automaton. However, BDDs are the most investigated data structures with a tool support unrivaled by other emerging representations. There are many highly optimized open-source BDD packages (e.g., Somenzi, 1996; Lind-Nielsen, 2001) that allow easy and efficient manipulation of BDDs. In contrast, publicly available, open-source compilers are still being developed for many newer representations. In particular, the application of BDDs to configuration resulted in a patent approval (Lichtenberg, Andersen, Hulgaard, Møller, & Rasmussen, 2001) and the establishment of the spinoff company Configit A/S[1].

The work in this paper is motivated by decision making scenarios where solutions are associated with a *cost function*, expressing implicitly properties such as price, quality, failure probability etc. A user might *prefer* one solution over another given the value of such properties. A natural way in which a user expresses his cost preferences in a configuration setting is to *bound* the minimal or maximal cost of any solution he is willing to accept. We therefore study the problem of calculating *weighted valid domains* (wCVD), where we eliminate those values that in every valid solution are more expensive than a user-provided maximal cost. We present a configurator that supports efficient cost bounding for a wide class of *additive* cost functions. Our approach is easily implementable and scales well for all the instances that were previously compiled into BDDs for standard interactive configuration. The cornerstone of our approach is to reuse the robust compilation of constraints into a BDD, and then *extract* a corresponding *multi-valued decision diagram* (MDD). The resulting MDD allows us to *label edges* with weights and utilize efficient shortest path algorithms to *label nodes* and filter expensive values on MDD edges. While our MDD extraction technique is novel, labeling edges in a decision diagram is suggested in other works as well. In its most generic interpretation (Wilson, 2005), edges of a decision diagram can be labeled with elements of a semiring to support algebraic computations relevant for probabilistic rea-soning, optimization etc. Amilhastre et al. (2002) suggest labeling edges of an automaton to reason abut optimal restorations and explanations. In general, many knowledge compilation structures have their weighted counterparts, many of which are captured in the framework of valued negation normal forms (VNNFs) (Fargier & Marquis, 2007). These structures are utilized for probabilistic reasoning, diagnosis, and other tasks involving reasoning about real-valued rather than Boolean functions. Some of them can in principle be used for `wCVD` queries, but the public tool support for weighted variants is less available or is tailored for tasks outside the configuration domain.

---

1. http://www.configit.com

We further extend our approach to support valid domains computation in the presence of *multiple cost functions*. A user often has multiple conflicting objectives, that should be satisfied simultaneously. Traditional approaches in multi-criteria optimization (Figueira, Greco, & Ehrgott, 2005; Ehrgott & Gandibleux, 2000) typically interact with a user in a way that is unsuitable in a configuration setting — cost functions are combined in a single objective and in each interaction step few non-dominated solutions are sampled and displayed to a user. Based on user selections a more adequate aggregation of costs is performed before the next interaction step. We suggest a more configuration-oriented interaction approach where domains are bounded with respect to multiple costs. We prove that this is a particularly challenging problem. Computing valid domains over an MDD in the presence of two cost functions (`2-wCVD`) is NP-hard, even in the simplest extension of linear inequalities with positive coefficients and Boolean variables. Despite this negative result, we provide an implementation of `2-wCVD` queries in *pseudo-polynomial* time and space and develop a *fully polynomial time approximation scheme* (FPTAS). We prove that no pseudo-polynomial algorithm and hence no fully polynomial approximation scheme exists for computing domains in the presence of arbitrarily many cost functions since that is an NP-hard problem in the strong sense. Finally, we demonstrate through experimental evaluation the applicability of both the `wCVD` and `2-wCVD` query over large real-world configuration models and product-catalogue datasets. To the best of our knowledge, we present the first interactive configurator supporting configuration wrt. cost restrictions in a backtrack-free and complete manner. This constitutes a novel addition to both existing product-configuration approaches as well as to approaches within multi-criteria decision making (Figueira et al., 2005).

The remainder of the paper is organized as follows. In Section 2 we describe background work and notation. In Section 3 we describe our approach to implementing `wCVD` query over an MDD while in Section 4 we show how to compile such an MDD. In Section 5 we discuss configuring in the presence of multiple costs. In Section 6 we present empirical evaluation of our approach. In Section 7 we describe related work and finally we conclude in Section 8.

## 2. Preliminaries

We will briefly review the most important concepts and background.

### 2.1 Constraint Satisfaction Problems

*Constraint satisfaction problems* (CSPs) form a framework for modeling and solving combinatorial problems, where a solution to a problem can be formulated as an assignment to variables that satisfy certain constraints. In its standard form, CSP involves only a finite number of variables, defined over finite domains.

**Definition 1 (CSP)** *A constraint satisfaction problem (CSP) is a triple $(X, D, F)$ where $X$ is a set of variables $\{x_1, \ldots, x_n\}$, $D = D_1 \times \ldots \times D_n$ is the Cartesian product of their finite domains $D_1, \ldots, D_n$ and $F = \{f_1, \ldots, f_m\}$ is a set of constraints defined over variables $X$. Each constraint $f$ is a function defined over a subset of variables $X_f \subseteq X$ called the* scope *of $f$. It maps each assignment to the $X_f$ variables into $\{0, 1\}$ where 1 indicates that $f$ is satisfied and 0 indicates that $f$ is violated by the assignment. The* solution *is an assignment to all variables $X$ that satisfies all constraints simultaneously.*

Formally, an *assignment* of values $a_1, \ldots, a_n$ to variables $x_1, \ldots, x_n$ is denoted as a set of pairs $\rho = \{(x_1, a_1), \ldots, (x_n, a_n)\}$. The domain of an assignment $dom(\rho)$ is the set of variables which are assigned: $dom(\rho) = \{x_i \mid \exists a \in D_i.(x_i, a) \in \rho\}$ and if all variables are assigned, i.e. $dom(\rho) = X$, we refer to $\rho$ as a *total assignment*. We say that a total assignment $\rho$ is *valid* if it satisfies all the rules, which is denoted as $\rho \models F$. A partial assignment $\rho, dom(\rho) \subseteq X$ is *valid* if it can be extended to a total assignment $\rho' \supseteq \rho$ that is valid $\rho' \models F$. We define the *solution space Sol* as the set of all valid total assignments, i.e. $Sol = \{\rho \mid \rho \models F, dom(\rho) = X\}$.

## 2.2 Interactive Configuration

*Interactive configuration* is an important application domain where a user is assisted in specifying a valid configuration (of a product, a service or something else) by interactively providing feedback on valid options for unspecified attributes. Such a problem arises in a number of domains. For example, when buying a product, a user should specify a number of product attributes. Some attribute combinations might not be feasible and if no guidance is provided, the user might reach a dead-end when interacting with the system. He will be forced to backtrack, which might seriously decrease the user satisfaction.

In many cases, valid configurations can be implicitly described by specifying restrictions on combining product attributes. We use a CSP model to represent such restrictions, and each CSP solution corresponds to a valid configuration. Each configurable attribute is represented with a variable, so that each attribute option corresponds to a value in the variable domain. In Example 1 we illustrate a simple configuration problem and its CSP model.

**Example 1** *To specify a T-shirt we have to choose the color (black, white, red, or blue), the size (small, medium, or large) and the print ("Men In Black" - MIB or "Save The Whales" - STW). If we choose the MIB print then the color black has to be chosen as well, and if we choose the small size then the STW print (including a large picture of a whale) cannot be selected as the picture of a whale does not fit on the small shirt. The configuration problem $(X, D, F)$ of the T-shirt example consists of variables $X = \{x_1, x_2, x_3\}$ representing color, size and print. Variable domains are $D_1 = \{0, 1, 2, 3\}$ (black, white, red, blue), $D_2 = \{0, 1, 2\}$ (small, medium, large), and $D_3 = \{0, 1\}$ (MIB, STW). The two rules translate to $F = \{f_1, f_2\}$, where $f_1$ is $x_3 = 0 \Rightarrow x_1 = 0$ (MIB $\Rightarrow$ black) and $f_2$ is $(x_2 = 0 \Rightarrow x_3 \neq 1)$ (small $\Rightarrow$ not STW). There are $|D_1||D_2||D_3| = 24$ possible assignments. Eleven of these assignments are valid configurations and they form the solution space shown in Fig. 1.* ◇

| | | |
|---|---|---|
| $(black, small, MIB)$ | $(black, large, STW)$ | $(red, large, STW)$ |
| $(black, medium, MIB)$ | $(white, medium, STW)$ | $(blue, medium, STW)$ |
| $(black, medium, STW)$ | $(white, large, STW)$ | $(blue, large, STW)$ |
| $(black, large, MIB)$ | $(red, medium, STW)$ | |

Figure 1: Solution space for the T-shirt example.

The fundamental task that we are concerned with in this paper is *calculating valid domains* (CVD) query. For a partial assignment $\rho$ representing previously made user assignments, the configurator calculates and displays a *valid domain* $VD_i[\rho] \subseteq D_i$ for each unassigned variable $x_i \in X \setminus dom(\rho)$. A domain is *valid* if it contains those and only those values with which $\rho$ can be extended to a total valid assignment $\rho'$. In our example, if a user selects a small T-shirt ($x_2 = 0$), valid domains should be restricted to a *MIB* print $VD_3 = \{0\}$ and *black* color $VD_1 = \{0\}$.

**Definition 2 (CVD)** *Given a CSP model $(X, D, F)$, for a given partial assignment $\rho$ compute valid domains:*

$$VD_i[\rho] = \{a \in D_i \mid \exists \rho'.(\rho' \models F \text{ and } \rho \cup \{(x_i, a)\} \subseteq \rho')\}$$

This task is of main interest since it delivers important interaction requirements: *backtrack-freeness* (user should never be forced to backtrack) and *completeness* (all valid configurations should be reachable) (Hadzic et al., 2004). There are other queries relevant for supporting user interaction such as explanations and restorations from a failure, recommendations of relevant products, etc., but CVD is an essential operation in our mode of interaction and is of primary importance in this paper.

## 2.3 Decision Diagrams

Decision diagrams form a family of rooted directed acyclic graphs (DAGs) where each node $u$ is labeled with a variable $x_i$ and each of its outgoing edges $e$ is labeled with a value $a \in D_i$. No node may have more than one outgoing edge with the same label. The decision diagram contains one or more *terminal* nodes, each labeled with a constant and having no outgoing edges. The most well known member of this family are *binary decision diagrams* (BDDs) (Bryant, 1986) which are used for manipulating Boolean functions in many areas, such as verification, model checking, VLSI design (Meinel & Theobald, 1998; Wegener, 2000; Drechsler, 2001) etc. In this paper we will primarily operate with the following variant of *multi-valued decision diagrams*:

**Definition 3 (MDD)** *An MDD denoted $M$ is a rooted directed acyclic graph $(V, E)$, where $V$ is a set of vertices containing the special terminal vertex $\mathbf{1}$ and a root $r \in V$. Further, $var: V \to \{1, \ldots, n+1\}$ is a labeling of all nodes with a variable index such that $var(\mathbf{1}) = n+1$. Each edge $e \in E$ is denoted with a triple $(u, u', a)$ of its start node $u$, its end node $u'$ and an associated value $a$.*

We work only with *ordered* MDDs. A total ordering $<$ of the variables is assumed such that for all edges $(u, u', a)$, $var(u) < var(u')$. For convenience we assume that the variables in $X$ are ordered according to their indices. Ordered MDDs can be considered as being arranged in $n$ *layers* of vertices, each layer being labeled with the same variable index. We will denote with $V_i$ the set of all nodes labeled with $x_i$, $V_i = \{u \in V \mid var(u) = i\}$. Similarly, we will denote with $E_i$ the set of all edges originating in $V_i$, i.e. $E_i = \{e(u, u', a) \in E \mid var(u) = i\}$. Unless otherwise specified, we assume that on each path from the root to the terminal, every variable labels exactly one node.

An MDD encodes a CSP solution set $Sol \subseteq D_1 \times \ldots \times D_n$, defined over variables $\{x_1, \ldots, x_n\}$. To check whether an assignment $\mathbf{a} = (a_1, \ldots, a_n) \in D_1 \times \ldots \times D_n$ is in $Sol$ we traverse $M$ from the root, and at every node $u$ labeled with variable $x_i$, we follow an edge labeled with $a_i$. If there is no such edge then $\mathbf{a}$ is not a solution, i.e., $\mathbf{a} \notin Sol$. Otherwise, if the traversal eventually ends in terminal $\mathbf{1}$ then $\mathbf{a} \in Sol$. We will denote with $p : u_1 \leadsto u_2$ any path in MDD from $u_1$ to $u_2$. Also, edges between $u$ and $u'$ will be sometimes denoted as $e : u \to u'$. A value $a$ of an edge $e(u, u', a)$ will be sometimes denoted as $v(e)$. We will not make distinction between paths and assignments. Hence, the set of all solutions represented by the MDD is $Sol = \{p \mid p : r \leadsto \mathbf{1}\}$. In fact, every node $u \in V_i$ can be associated with a subset of solutions $Sol(u) = \{p \mid p : u \leadsto \mathbf{1}\} \subseteq D_i \times \ldots \times D_n$.

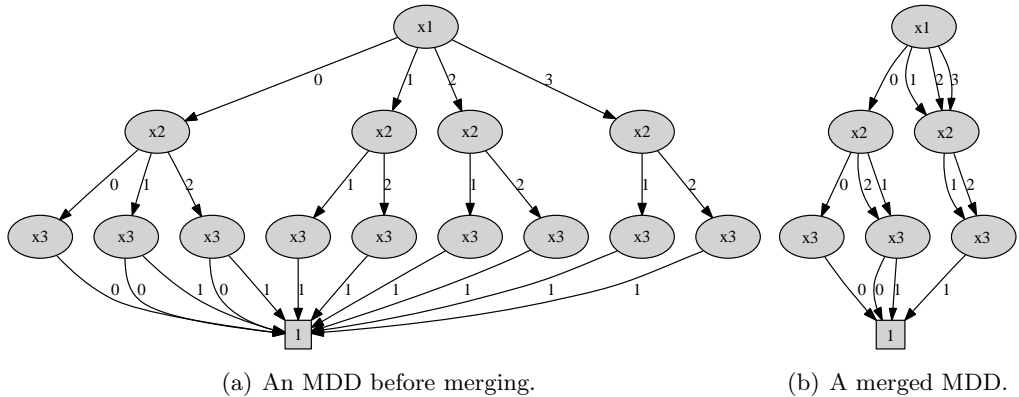

(a) An MDD before merging.  (b) A merged MDD.

Figure 2: An uncompressed and merged MDD for the T-Shirt example.

Decision diagrams can be exponentially smaller than the size of the solution set they encode by *merging isomorphic subgraphs*. Two nodes $u_1, u_2$ are isomorphic if they encode the same solution set $Sol(u_1) = Sol(u_2)$. In Figure 2 we show a fully expanded MDD 2(a) and an equivalent merged MDD 2(b) for the T-shirt solution space. In addition to merging isomorphic subgraphs, another compression rule is usually utilized: *removing redundant nodes*. A node $u \in V_i$ is *redundant* if it has $D_i$ outgoing edges, each pointing to the same node $u'$. Such nodes are *eliminated* by redirecting incoming edges from $u$ to $u'$ and deleting $u$ from $V$. This introduces *long edges* that skip layers. An edge $e(u, u', a)$ is long if $var(u)+1 < var(u')$. In this case, $e$ encodes the set of solutions: $\{a\} \times D_{var(u)+1} \times \ldots \times D_{var(u')-1}$. We will refer to an MDD where both merging of isomorphic nodes and removal of redundant nodes have taken place as a *reduced MDD*, which constitutes a multi-valued generalization of BDDs which are typically reduced and ordered. A reduced MDD for the T-shirt CSP is shown in Figure 3. In this paper, unless emphasized otherwise, by MDD we always assume an ordered *merged* but not reduced MDD, since exposition is simpler, and removal of redundant nodes can have at most a linear effect on size. Given a variable ordering there is a unique merged MDD for a given CSP $(X, D, F)$ and its solution set *Sol*. The size of MDD depends critically on the ordering, and could vary exponentially. It can grow exponentially with the number of variables, but in practice, for many interesting problems the size is surprisingly small.

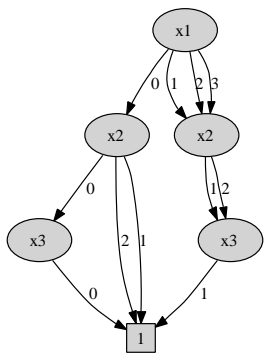

Figure 3: A reduced MDD for the T-shirt example.

**Interactive Configuration over Decision Diagrams.** A particularly attractive property of decision diagrams is that they support efficient execution of a number of important queries, such as checking for consistency, validity, equivalence, counting, optimization etc. This is utilized in a number of application domains where most of the problem description is known offline (diagnosis, verification,etc.). In particular, calculating valid domains is linear in the size of the MDD. Since calculating valid domains is an NP-hard problem in the size of the input CSP model, it is not possible to guarantee interactive response in *real-time*. In fact, the unacceptably long worst-case response times have been empirically observed in a purely search-based approach to computing valid domains (Subbarayan et al., 2004). Therefore, by compiling CSP solutions *off-line* (prior to user interaction) into a decision diagram, we can efficiently (in the size of the MDD) compute valid domains during online interaction with a user. It is important to note that the order in which the user decides variables is completely unconstrained, i.e. it does not depend on the ordering of MDD variables. In our previous work we utilized *Binary Decision Diagrams* (BDDs) to represent all valid configurations so that CVD queries can be executed efficiently (Hadzic et al., 2004). Of course, BDDs might be exponentially large in the input CSP, but for many classes of constraints they are surprisingly compact.

## 3. Interactive Cost Processing over MDDs

The main motivation for this work is extending the interactive configuration approach of Møller et al. (2002), Hadzic et al. (2004), Subbarayan et al. (2004) to situations where in addition to a CSP model $(X, D, F)$ involving only hard constraints, there is also a *cost function*:

$$c : D_1 \times \ldots \times D_n \to \mathbb{R}.$$

In *product configuration* setting, this could be a product price. In *uncertainty* setting, the cost function might indicate a probability of an occurrence of an event represented by a solution (failure of a hardware component, withdrawal of a bid in an auction etc.). In any *decision support* context, the cost function might indicate user preferences. There is a number of cost-related queries in which a user might be interested, e.g. finding an optimal solution, or computing a most probable explanation. We, however, assume that a user is interested in tight control of both the variable values as well as the cost of selected solutions. For example, a user might desire a specific option $x_i = a$, but he would also care about how would such an assignment affect the cost of the remaining optimal solutions. We should communicate this information to the user, and allow him to strike the right balance between the cost and variable values by allowing him to interactively limit the maximal cost of the product in addition to assigning variable values. Therefore, in this paper we are primarily concerned with implementing a *weighted CVD* (`wCVD`) query: for a user-specified maximum cost $K$, we should indicate which values in the unassigned variable domains can be extended to a total assignment that is valid *and* costs less than $K$. From now on, we assume that a user is interested in bounding the maximal cost (limiting the minimal cost is symmetric).

**Definition 4 (`wCVD`)** *Given a CSP model $(X, D, F)$, a cost function $c : D \rightarrow \mathbb{R}$ and a maximal cost $K$, for a given partial assignment $\rho$ a weighted CVD (`wCVD`) query requires computation of the valid domains:*

$$VD_i[\rho, K] = \{a \in D_i \mid \exists \rho'.(\rho' \models F \text{ and } \rho \cup \{(x_i, a)\} \subseteq \rho' \text{ and } c(\rho') \leq K)\}$$

In this section we assume that an MDD representation of all CSP solutions is already generated in an offline compilation step. We postpone discussion of MDD compilation to Section 4 and discuss only delivering efficient online interaction on top of such MDD. We will first discuss the practicability of implementing `wCVD` queries through explicit encoding of costs into an MDD. We will then provide a practical and efficient approach to implementing `wCVD` over an MDD when the cost function is *additive*. Finally, we will discuss further extensions to handling more expressive cost functions.

### 3.1 Handling Costs Explicitly

An immediate approach to interactively handling a cost function is to treat the cost as any other solution attribute, i.e. to add a variable $y$ to variables $X$ and add the constraint

$$y = c(x_1, \ldots, x_n) \tag{1}$$

to formulas $F$ to enforce that $y$ is equal to the total cost. The resulting configuration model is compiled into an MDD $M'$ and a user is able to bound the cost by restricting the domain of $y$.

Assuming the variable ordering $x_1 < \ldots < x_n$ in the original CSP model $(X, D, F)$, and assuming we inserted a cost variable into the $i$-th position, the new variable set $X'$ has a variable ordering $x'_1 < \ldots < x'_{n+1}$ s.t. $x'_1 = x_1, \ldots, x'_{i-1} = x_{i-1}, x'_i = y$ and $x'_{i+1} = x_i, \ldots, x'_{n+1} = x_n$. The domain $D'_i$ of variable $x'_i$ is the set of all feasible costs $C(Sol) = \{c(s) \mid s \in Sol\}$. We will now demonstrate that the MDD $M'$ may be exponentially larger than $M$.

**Lemma 1** $|E'_i| \geq |C(Sol)|$.

**Proof 1** *For the $i$-th layer of MDD $M'$ corresponding to variable $y$, for each cost $c \in C(Sol)$ there must be at least one path $p : r \rightsquigarrow 1$ with $c(p) = c$, and for such a path, an edge $e \in E'_i$ at the $i$-th layer must be labeled with $v(e) = c$. Hence, for each cost there must be at least one edge in $E'_i$. This proves the lemma.*

Furthermore, at least one of the layers of nodes $V'_i, V'_{i+1}$ has a number of nodes greater than $\sqrt{|E'_i|}$. This follows from the following lemma:

**Lemma 2** *For the $i$-th layer of MDD $M'$, $|V'_i| \cdot |V'_{i+1}| \geq |E'_i|$.*

**Proof 2** *Since there are at most $|V'_i| \cdot |V'_{i+1}|$ pairs of nodes $(u_1, u_2) \in V'_i \times V'_{i+1}$, the statement follows from the fact that for each pair $(u_1, u_2)$ there can be at most one edge $e : u_1 \rightarrow u_2$. Namely, every solution $p_3$ formed by concatenating paths $p_1 : r \rightsquigarrow u_1$ and $p_2 : u_2 \rightsquigarrow 1$ has a unique cost $c(p_3)$. However, if there were two edges $e_1, e_2 : u_1 \rightarrow u_2$, they would have to have different values $v(e_1) \neq v(e_2)$. But then, the same solution $c(p_3)$ would correspond to two different costs $v(e_1), v(e_2)$.*

From the above considerations we see that whenever the range of possible costs $C(Sol)$ is exponential, the resulting MDD $M'$ would be exponentially large as well. This would result in a significantly increased size $|V'|/|V|$, particularly when there is a large number of isomorphic nodes in $M$ that would become non-isomorphic once the variable $y$ is introduced (since they root paths with different costs). An extreme instance of such a behavior is presented in Example 2. Furthermore, even if $C(Sol)$ is not large, there could be orders of magnitude of increase in the size of $M'$ due to breaking of isomorphic nodes in the MDD as will be empirically demonstrated in Section 6, Table 3, for a number of configuration instances. This is a major disadvantage as otherwise efficient *CVD* algorithms become unusable since they operate over a significantly larger structure.

**Example 2** *Consider a model $C(X, D, F)$ with no constraints $F = \{\}$, and Boolean variables $D_j = \{0, 1\}$, $j = 1 \ldots, n$. The solution space includes all assignments $Sol = D_1 \times \ldots \times D_n$ and a corresponding MDD $M(V, E)$ has one vertex and two edges at each layer, $|V| = n + 1$, $|E| = 2 \cdot n$. If we use the cost function: $c(x_1, \ldots, x_n) = \sum_{j=1}^{n} 2^{j-1} \cdot x_j$, there is an exponential number of feasible costs $C(Sol) = \{0, \ldots, 2^n - 1\}$. Hence, $|E'_i| \geq 2^n$ and for the $i$-th layer corresponding to variable $y$, at least one of the layers $|V'_i|, |V'_{i+1}|$ is greater than $\sqrt{2^n} = 2^{n/2}$.*

However, if there was no significant node isomorphism in $M$, adding a $y$ variable does not necessarily lead to a significant increase in size. An extreme instance of this is an MDD with no isomorphic nodes, for example when every edge is labeled with a unique value. For such an MDD, the number of non-terminal nodes is $n \cdot |Sol|$. By adding a cost variable $y$, the resulting MDD would add at most one node per path, leading to an MDD with at most $(n + 1) \cdot |Sol|$ nodes. This translates to a minor increase in size: $|V'|/|V| = (n + 1)/n$. This property will be empirically demonstrated in Section 6, Table 3, for product-catalogue datasets. In the remainder of this paper we develop techniques tailored for instances where a large increase in size occurs. We avoid explicit cost encoding and aim to exploit the structure of the cost function to implement `wCVD`.

## 3.2 Processing Additive Cost Functions

One of the main contributions of this paper is a practical and efficient approach to deliver `wCVD` queries if the cost function is *additive*. An additive cost function has the form

$$c(x_1, \dots, x_n) = \sum_{i=1}^n c_i(x_i)$$

where a cost $c_i(a_i) \in \mathbb{R}$ is assigned for every variable $x_i$ and every value in its domain $a_i \in D_i$.

Additive functions are one of the most important and frequently used modeling constructs. A number of important combinatorial problems are modeled as *integer linear programs* where often both the constraints and the objective function are linear, i.e. represent special cases of additive cost functions. In *multi-attribute utility theory* user preferences are under certain assumptions aggregated into a single additive function through weighted summation of utilities of individual attributes. In a product configuration context, many properties are additive such as the memory capacity of a computer or the total weight. In particular, based on our experience in commercially applying configuration technology, the price of a product can often be modeled as the (weighted) sum of prices of individual parts.

### 3.2.1 THE LABELING APPROACH

Assuming that we are given an MDD representation of the solution space *Sol* and a cost function $c$, our approach to answering `wCVD` queries is based on three steps: 1) restricting MDD wrt. the latest user assignment, 2) labeling remaining nodes by executing shortest path algorithms and 3) filtering too expensive values by using node labels.

**Restricting MDD.** We are given a user assignment $x_i = a_i$, where $x_i$ can be any of the unassigned variables, regardless of its position in the MDD variable ordering. We initialize MDD pruning by removing all edges $e(u, u', a)$, that are not in agreement with the latest assignment, i.e. where $var(u) = i$ and $a \neq a_i$. This might cause a number of other edges and nodes to become unreachable from the terminal or the root if we removed the last edge in the set of children edges $Ch(u)$ or parent edges $P(u')$. Any unreachable edge must be removed as well. The pruning is repeated until a fixpoint is reached, i.e. until no more nodes or edges can be removed. Algorithm 1 implements this scheme in $O(|V| + |E|)$ time and space by using a queue $Q$ to maintain the set of edges that are yet to be removed.

Note that *unassigning* a user assignment $x_i = a_i$ can be easily implemented in linear time as well. It suffices to restore a copy of the initial MDD $M$, and perform restriction wrt. a partial assignment $\rho \setminus \{(x_i, a_i)\}$ where $\rho$ is a current assignment. Algorithm 1 is easily extended for this purpose by initializing the edge removal list $Q$ with edges incompatible wrt. *any* of the assignments in $\rho$.

**Computing Node Labels.** Remaining edges $e(u, u', a)$ in each layer $E_i$ are implicitly labeled with $c(e) = c_i(a)$. In the second step we compute for each MDD node $u \in V$ an *upstream* cost of the shortest path from the root $r$ to $u$, denoted as $U[u]$, and a *downstream* cost of the shortest path from $u$ to the terminal $\mathbf{1}$, denoted as $D[u]$:

$$U[u] = \min_{p:r \rightsquigarrow u} \left\{ \sum_{e \in p} c(e) \right\}, D[u] = \min_{p:u \rightsquigarrow \mathbf{1}} \left\{ \sum_{e \in p} c(e) \right\} \tag{2}$$

---

**Algorithm 1**: Restrict MDD.

**Data**: MDD $M(V, E)$, variable $x_i$, value $a_i$
**foreach** $e \in E_i, v(e) \neq a_i$ **do**
    $Q.push(e)$;
**while** $Q \neq \emptyset$ **do**
    $e(u, u', a) \leftarrow Q.pop()$;
    delete $e$ from $M$;
    **if** $Ch(u) = \emptyset$ **then**
        **foreach** $e : u'' \rightarrow u$ **do**
            $Q.push(e)$;
    **if** $P(u') = \emptyset$ **then**
        **foreach** $e : u' \rightarrow u''$ **do**
            $Q.push(e)$;

---

Algorithm 2 computes $U[u]$ and $D[u]$ labels in $\Theta(|V| + |E|)$ time and space.

---

**Algorithm 2**: Update $U, D$ labels.

**Data**: MDD $M(V, E)$, Cost function $c$
$D[\cdot] = \infty$, $D[\mathbf{1}] = 0$;
**foreach** $i = n, \ldots, 1$ **do**
    **foreach** $u \in V_i$ **do**
        **foreach** $e : u \rightarrow u'$ **do**
            $D[u] = min\{D[u], c(e) + D[u']\}$
$U[\cdot] = \infty$, $U[r] = 0$;
**foreach** $i = 1, \ldots, n$ **do**
    **foreach** $u \in V_i$ **do**
        **foreach** $e : u \rightarrow u'$ **do**
            $U[u'] = min\{U[u'], c(e) + U[u]\}$

---

**Computing Valid Domains.** Once the upstream and downstream costs $U, D$ are computed, we can efficiently compute valid domains $VD_i$ wrt. any maximal cost bound $K$ since:

$$VD_i[K] = \{v(e) \mid U[u] + c(e) + D[u'] \leq K, \ e : u \rightarrow u', \ u \in V_i\} \tag{3}$$

This can be achieved in a linear-time traversal $\Theta(|V| + |E|)$ as shown in Algorithm 3.

---

**Algorithm 3**: Compute valid domains.

**Data**: MDD $M(V, E)$, Cost function $c$, Maximal cost $K$
**foreach** $i = 1, \ldots, n$ **do**
    $VD_i = \emptyset$;
    **foreach** $u \in V_i$ **do**
        **foreach** $e : u \rightarrow u'$ **do**
            **if** $U[u] + c[e] + D[u'] \leq K$ **then**
                $VD_i \leftarrow VD_i \cup \{v(e)\}$;

---

Hence the overall interaction is as follows. Given a current partial assignment $\rho$, MDD is restricted wrt. $\rho$ through Algorithm 1. Labels $U, D$ are then computed through Algorithm 2 and valid domains are computed using Algorithm 3. The execution of all of these algorithms requires $\Theta(|V| + |E|)$ time and space. Hence, when an MDD representation of the solution space is available, we can interactively enforce additive cost restrictions in linear time and space.

### 3.3 Processing Additive Costs Over Long Edges

Our scheme can be extended to MDDs containing long edges. While for multivalued CSP models with large domains space savings due to long edges might not be significant, for binary models and *binary decision diagrams* (BDDs) more significant savings are possible. Furthermore, in a similar fashion, our scheme might be adopted over other versions of decision diagrams that contain long edges (with different semantics) such as *zero-suppressed BDDs* where a long edge implies that all skipped variables are assigned 0.

Recall that in reduced MDDs, *redundant* nodes $u \in V_i$ which have $D_i$ outgoing edges, each pointing to the same node $u'$, are *eliminated*. An edge $e(u, u', a)$ with $var(u) = k$ and $var(u') = l$ is long if $k + 1 < l$, and in this case, $e$ encodes a set of solutions: $\{a\} \times D_{k+1} \times \ldots \times D_{l-1}$. The labeling of edges can be generalized to accommodate such edges as well. Let domains $D'_j$, $j = 1, \ldots, n$ represent variable domains updated wrt. the current assignment, i.e. $D'_j = D_j$ if $x_j$ is unassigned, and $D'_j = \{\rho[x_j]\}$ otherwise. An edge $e(u, u', a)$, $(var(u) = k, var(u') = l)$ is removed if $a \notin D'_k$ in an analogous way to the MDD pruning in the previous subsection. Otherwise, it is labeled with

$$c(e) = c_k(a) + \sum_{j=k+1}^{l-1} \min_{a' \in D'_j} c_j(a') \qquad (4)$$

which is the cost of the cheapest assignment to $x_k, \ldots, x_{l-1}$ consistent with the edge and the partial assignment $\rho$. Once the edges are labeled, the upstream and downstream costs $U, D$ are computed in $\Theta(|V| + |E|)$ time, in the same manner as in the previous subsection.

However, computing valid domains has to be extended. As before, a *sufficient* condition for $a \in VD_i$ is the existence of an edge $e : u \to u'$, originating in the $i$-th layer $u \in V_i$ such that $v(e) = a$ and

$$U[u] + c[e] + D[u'] \leq K. \qquad (5)$$

However, this is no longer a *necessary* condition, as even if there is no edge satisfying (5), there could exist a long edge *skipping* the $i$-th layer that still allows $a \in VD_i$. We therefore, for each layer $i$, have to compute the cost of the cheapest path skipping the layer:

$$P[i] = \min\{U[u] + c(e) + D[u'] \mid e : u \to u' \in E, \ var(u) < i < var(u')\} \qquad (6)$$

If there is no edge skipping the $i$-th layer, we set $P[i] = \infty$. Let $c_{min}[i]$ denote the cheapest value in $D'_i$, i.e. $c_{min}[i] = \min_{a \in D'_i} c_i(a)$. To determine if there is a *long* edge allowing $a \in VD_i$, for an unassigned variable $x_i$, the following must hold:

$$P[i] + c_i(a) - c_{min}[i] \leq K \qquad (7)$$

Finally, a sufficient and necessary condition for $a \in VD_i$ is that one of the conditions (5) and (7) holds. If variable $x_i$ is assigned with a value drawn from a valid domain in a previous step, we are guaranteed that $VD_i = \{\rho[x_i]\}$ and no calculations are necessary. Labels $P[i]$

---

**Algorithm 4**: Update $P$ labels.

---

**Data**: MDD $M(V, E)$, Cost function $c$

$P[\cdot] = \infty$;

**foreach** $i = 1, \ldots, n$ **do**
    **foreach** $u \in V_i$ **do**
        **foreach** $e : u \to u'$ **do**
            **foreach** $j \in \{var(u) + 1, \ldots, var(u') - 1\}$ **do**
                $P[j] = min\{P[j], U[u] + c(e) + D[u']\}$;

---

can be computed by Algorithm 4 in worst-case $O(|E| \cdot n)$ time. Note that this bound is over-pessimistic as it assumes that every edge in $|E|$ is skipping every variable in $X$.

Once the auxiliary structures $U, D, P$ are computed, valid domains can be efficiently extracted using Algorithm 5. For each unassigned variable $x_i$, value $a \in D_i$ is in a valid domain $VD_i[K]$ iff the following holds: condition (7) is satisfied or for an edge $e(u, u', a) \in E$ condition (5) is satisfied. For each non-assigned variable $i$, the algorithm first checks for each value $a \in D_i$ whether it is supported by a skipping edge $P[i]$. Afterwards, it scans the $i$-th layer and extracts values supported by edges $E_i$. This is achieved in $\Theta(|D| + |V| + |E|)$ time, where $|D| = \sum_{i=1}^{n} |D_i|$.

---

**Algorithm 5**: Computing valid domains $VD_i$.

---

**Data**: MDD $M(V, E)$, cost function $C$, maximal cost $K$

**foreach** $i = 1, \ldots, n$ **do**
    $VD_i = \emptyset$;
    **if** $x_i$ *assigned to* $a_i$ **then**
        $VD_i \leftarrow \{a_i\}$;
        continue;
    **foreach** $a \in D_i$ **do**
        **if** $P[i] + c_i(a) - c_{min}[i] \leq K$ **then**
            $VD_i \leftarrow VD_i \cup \{a\}$;
    **foreach** $u \in V_i$ **do**
        **foreach** $e : u \to u'$ **do**
            **if** $U[u] + c[e] + D[u'] \leq K$ **then**
                $VD_i \leftarrow VD_i \cup \{v(e)\}$;

---

Again, the overall interaction remains the same. Labels $P$ can be incrementally updated in worst case $O(|E| \cdot n)$ time. Valid domains are then extracted in $\Theta(|D| + |V| + |E|)$ time. In response to changing a cost restriction $K$, auxiliary labels need not be updated. Valid domains are extracted directly using Algorithm 5 in $\Theta(|D| + |V| + |E|)$ time.

### 3.4 Handling Non-Additive Cost Functions

In certain interaction settings, the cost function is not additive. For example, user preferences might depend on an entire package of features rather than a selection of each individual feature. Similarly, the price of a product need not be a simple sum of costs of individual parts, but might depend on combinations of parts that are selected. In general, our cost function $c(x_1, \ldots, x_n)$ might be a sum of non-unary cost functions $c_i$, $i = 1, \ldots, k$,

$$c(x_1, \ldots, x_n) = \sum_{i=1}^{k} c_i(X_i)$$

where each cost function $c_i$ expresses a unique contribution of combination of features within a subset of variables $X_i \subseteq X$,

$$c_i : \prod_{j \in X_i} D_j \to \mathbb{R}.$$

### 3.4.1 NON-UNARY LABELING

Our approach can be extended to handle non-unary costs by adopting labeling techniques that are used with other graphical representations (e.g., Wilson, 2005; Mateescu et al., 2008). Assume we are given a cost function $c(x_1, \ldots, x_n) = \sum_{i=1}^{k} c_i(X_i)$. Let $A(i)$ denote the set of all cost functions $c_j$ such that $x_i$ is the last variable in the scope of $c_j$:

$$A(i) = \{c_j \mid x_i \in X_j \text{ and } x_{i'} \notin X_j, \forall i' > i\}.$$

Given assignment $\mathbf{a}(a_1, \ldots, a_i)$ to variables $x_1, \ldots, x_i$, we can evaluate every function $c_j \in A_i$. If the scope of $c_j$ is a strict subset of $\{x_1, \ldots, x_i\}$, we set $c_j(\mathbf{a})$ to be the value of $c_j(\pi_{X_j}(\mathbf{a}))$ where $\pi_{X_j}(\mathbf{a})$ is a projection of $\mathbf{a}$ onto $X_j$. Now, for every path $p : r \rightsquigarrow u$, $u \in V_{i+1}$, and its last edge (in the $i$-th layer) $e \in E_i$, we label $e$ with the sum of all cost functions that have become completely instantiated after assigning $x_i = a_i$:

$$c(e, p) = \sum_{c_j \in A(i)} c_j(p). \tag{8}$$

With respect to such labeling, a cost of a solution represented by a path $p$ would indeed be the sum of costs of its edges: $\sum_{e \in p} c(e, p)$. In order to apply our approach developed for additive cost functions in Section 3.2, each edge should be labeled with a cost that is the same for any incoming path. However, this is not possible in general. We therefore have to expand the original MDD, by creating multiple copies of $e$ and splitting incoming paths to ensure that any two paths $p_1, p_2$ sharing a copy $e'$ of an edge $e$ induce the same edge cost $c(e', p_1) = c(e', p_2)$. Such an MDD, denoted as $M_c$, can be generated using for example search with caching isomorphic nodes as suggested by Wilson (2005), or by extending the standard *apply* operator to handle weights as suggested by Mateescu et al. (2008).

### 3.4.2 IMPACT ON THE SIZE

The increase in size of $M_c$ relatively to the cost-oblivious version $M$ depends on the "additivity" of the cost function $c$. For example, for fully additive cost functions (each scope $X_i$ contains a single variable) $M_c = M$, since a label on $c(e)$ is the same regardless of the incoming path. However, if the entire cost function $c$ is a single non-additive component $c_1(X_1)$ with global scope ($X_1 = X$), then only the edges in the last MDD layer are labeled, as in the case of explicit cost encoding into MDD from Section 3.1. There must be at least $C(Sol)$ edges in the last layer, one for each feasible cost. Hence, if the range of costs $C(Sol)$ is exponential, so is the size of $M_c$. Furthermore, even if $C(Sol)$ is of limited size, an increase in $M_c$ might be significant due to breakup of node isomorphisms in previous layers. In case of explicit cost encoding (Section 3.1) such an effect is demonstrated empirically in Section 6. A similar effect on the size would occur in other graphical-representations. For example, in representations exploiting global CSP structure - such as weighted cluster trees (Pargamin, 2003) - adding non-additive cost functions increases the size of the clusters, as it is required that for each non-additive component $c_i(X_i)$ at least one cluster contains the entire scope $X_i$. Furthermore, criteria for node merging of Wilson (2005) and Mateescu et al. (2008) are more refined, since nodes are no longer isomorphic if they root the same set of feasible paths, but the paths must be of the same cost as well.

### 3.4.3 Semiring Costs and Probabilistic Queries

Note that our approach can be further generalized to accommodate more general aggregation of costs as discussed by Wilson (2005). Cost functions $c_i$ need not map assignments of $X_i$ variables into the set of real numbers $\mathbb{R}$ but to any set $A$ equipped with operators $\oplus, \otimes$ such that $\mathcal{A} = (A, \mathbf{0}, \mathbf{1}, \oplus, \otimes)$ is a *semiring*. The MDD property that is computed is $\oplus_{p:r \rightsquigarrow \mathbf{1}} \otimes_{e \in p} c(e)$. Operator $\otimes$ aggregates edge costs while operator $\oplus$ aggregates path costs. In a semiring $\oplus$ distributes over $\otimes$, and the global computation can be done efficiently by local node-based aggregations, much as a shortest path is computed. Our framework is based on reasoning about paths of minimal cost which corresponds to using $\mathcal{A} = (\mathbb{R}^+, 0, 1, min, +)$ but different semirings could be used. In particular, by taking $\mathcal{A} = (\mathbb{R}^+, 0, 1, +, \times)$ we can handle probabilistic reasoning. Each cost function $c_i$ corresponds to a *conditional probability table*, the cost of an edge $c(e)$, $e : u \rightarrow u' \in E_i$ corresponds to the probability of $P(x_i = v(e))$ given any of the assignments $p : r \rightsquigarrow u$. The cost of a path $c(p) = \prod_{e \in p} c(e)$ is a probability of an event represented by the path, and for a given value $a \in D_i$ we can get the *marginal probability* of $P(x_i = a)$ by computing $\sum_{e(u,u',a) \in E_i}(U[u] \times c(e) \times D[u'])$.

## 4. Compiling MDDs

In the previous section we showed how to implement cost queries once the solution space is represented as an MDD. In this section, we discuss how to generate such MDDs from a CSP model description $(X, D, F)$. Our goal is to develop an efficient and easy to implement approach that can handle all instances handled previously through BDD-based configuration (Hadzic et al., 2004).

**Variable Ordering.** The first step is to choose an ordering for CSP variables $X$. This is critical since different variable orders could lead to exponential differences in MDD size. This is a well investigated problem, especially for binary decision diagrams. For a fixed formula, deciding if there is an ordering such that the resulting BDD would have at most $T$ nodes (for some threshold $T$) is an NP-hard problem (Bollig & Wegener, 1996). However, there are well developed heuristics, that either exploit the structure of the input model or use variable swapping in existing BDD to improve the ordering in a local-search manner (Meinel & Theobald, 1998). For example, *fan-in* and *weight* heuristics are popular when the input is in the form of a combinational circuits. If the input is a CSP, a reasonable heuristic is to choose an ordering that minimizes the *path-width* of the corresponding *constraint graph*, as an MDD is in worst case exponential in the path-width (Bodlaender, 1993; Wilson, 2005; Mateescu et al., 2008). Investigating heuristics for variable ordering is out of the scope of our work, and in the remainder of this paper we assume that the ordering is already given. In all experiments we use default orderings provided for the instances.

**Compilation Technique.** Our approach is to first compile a CSP model into a binary decision diagrams (BDD) by exploiting highly optimized and stable BDD packages (e.g., Somenzi, 1996) and afterwards *extract* the corresponding MDD. Dedicated MDD packages are rare, provide limited functionality and their implementations are not as optimized as BDD packages to offer competitive performance (Miller & Drechsler, 2002). An interesting recent alternative is to generate BDDs through search with caching isomorphic nodes. Such an approach was suggested by Huang and Darwiche (2004) to compile BDDs from CNF formulas, and it proved to be a valuable addition to standard compilation based on pairwise BDD conjunctions. However, such compilation technology is still in the early stages of development and an open-source implementation is not publicly available.

## 4.1 BDD Encoding

Regardless of the BDD compilation method, the finite domain CSP variables $X$ first have to be *encoded* by Boolean variables. Choosing a proper encoding is important since the intermediate BDD might be too large or inadequate for subsequent extraction. In general, each CSP variable $x_i$ would be encoded with $k_i$ Boolean variables $\{x_1^i, \ldots, x_{k_i}^i\}$. Each $a \in D_i$ has to be mapped into a bit vector $enc_i(a) = (a_1, \ldots, a_{k_i}) \in \{0,1\}^{k_i}$ such that for different values $a \neq a'$ we get different vectors $enc_i(a) \neq enc_i(a')$. There are several standard Boolean encodings of multi-valued variables (Walsh, 2000). In the *log encoding* scheme each $x_i$ is encoded with $k_i = \lceil log|D_i| \rceil$ Boolean variables, each representing a digit in binary notation. A multivalued assignment $x_i = a$ is translated into a set of assignments $x_j^i = a_j$ such that $a = \sum_{j=1}^{k_i} 2^{j-1} a_j$. Additionally, a domain constraint $\sum_{j=1}^{k_i} 2^{j-1} x_j^i < |D_i|$ is added to forbid those bit assignments $(a_1^i, \ldots, a_{k_i}^i)$ that encode values outside domain $D_i$. The *direct encoding* (or *1-hot encoding*) is also common, and especially well suited for efficient propagation when searching for a single solution. In this scheme, each multi-valued variable $x_i$ is encoded with $|D_i|$ Boolean variables $\{x_1^i, \ldots, x_{k_i}^i\}$, where each variable $x_j^i$ indicates whether the $j$-th value in domain $a_j \in D_i$ is assigned. For each variable $x_i$, exactly one value from $D_i$ has to be assigned. Therefore, we enforce a domain constraint $x_1^i + \ldots + x_{k_i}^i = 1$ for each $i = 1, \ldots, n$. Hadzic, Hansen, and O'Sullivan (2008) have empirically demonstrated that using log encoding rather than direct encoding yields smaller BDDs.

The set of Boolean variables is fixed as the union of all encoding variables, $X_b = \bigcup_{i=1}^{n} \{x_1^i, \ldots, x_{k_i}^i\}$ but we still have to specify the ordering. A common ordering that is well suited for efficiently answering configuration queries is *clustered ordering*. Here, Boolean variables $\{x_1^i, \ldots, x_{k_i}^i\}$ are grouped into blocks that respect the ordering among finite-domain variables $x_1 < \ldots < x_n$. That is,

$$x_{j_1}^{i_1} < x_{j_2}^{i_2} \Leftrightarrow i_1 < i_2 \vee (i_1 = i_2 \wedge j_1 < j_2).$$

There might be other orderings that yield smaller BDDs for specific classes of constraints. Bartzis and Bultan (2003) have shown that linear arithmetic constraints can be represented more compactly if Boolean variables $x_j^i$ are grouped wrt. bit-position $j$ rather than the finite-domain variable $x_i$, i.e. $x_{j_1}^{i_1} < x_{j_2}^{i_2} \Leftrightarrow j_1 < j_2 \lor (j_1 = j_2 \land i_1 < i_2)$. However, configuration constraints involve not only linear arithmetic constraints, and space savings reported by Bartzis and Bultan (2003) are significant only when all the variable domains have a size that is a power of two. Furthermore, clustered orderings yield BDDs that preserve essentially the same combinatorial structure which allows us to extract MDDs efficiently as will be seen in Section 4.2.

**Example 3** *Recall that in the T-shirt example* $D_1 = \{0, 1, 2, 3\}$, $D_2 = \{0, 1, 2\}$, $D_3 = \{0, 1\}$. *The* log encoding *variables are* $x_1^1 < x_2^1 < x_1^2 < x_2^2 < x_1^3$, *inducing a variable set* $X_b = \{1, 2, 3, 4, 5\}$. *The log-BDD with clustered variable ordering is shown in Figure 4(a).* ◇

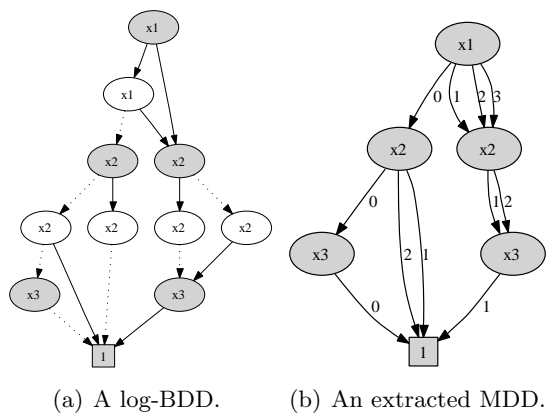

(a) A log-BDD.     (b) An extracted MDD.

Figure 4: A log-BDD with clustered ordering, and an extracted MDD for the T-shirt example. For BDD, we draw only the terminal node **1** while terminal node **0** and its incoming edges are omitted for clarity. Each node corresponding to a Boolean encoding variable $x_j^i$ is labeled with the corresponding CSP variable $x_i$. Edges labeled with 0 and 1 are drawn as dashed and full lines, respectively.

## 4.2 MDD Extraction

Once the BDD is generated using clustered variable ordering we can extract a corresponding MDD using a method which was originally suggested by Hadzic and Andersen (2006) and that was subsequently expanded by Hadzic et al. (2008). In the following considerations, we will use a mapping $cvar(x_j^i) = i$ to denote the CSP variable $x_i$ of an encoding variable $x_j^i$ and, with a slight abuse of notation, we will apply $cvar$ also to BDD nodes $u$ labeled with $x_j^i$. For terminal nodes, we define $cvar(\mathbf{0}) = cvar(\mathbf{1}) = n + 1$ (recall that BDD has two terminal nodes **0** and **1** indicating *false* and *true* respectively). Analogously, we will use a mapping $pos(x_j^i) = j$ to denote the position of a bit that the variable is encoding.

Our method is based on recognizing a subset of BDD nodes that captures the core of the MDD structure, and that can be used directly to construct the corresponding MDD.

In each block of BDD layers corresponding to a CSP variable $x_i$, $L_i = V_{x_1^i} \cup \ldots \cup V_{x_{k_i}^i}$, it suffices to consider only those nodes that are reachable by an edge from a previous block of layers:

$$In_i = \{u \in L_i \mid \exists_{(u',u) \in E} \; cvar(u') < cvar(u)\}.$$

For the first layer we take $In_1 = \{r\}$. The resulting MDD $M(V', E')$ $M$ contains only nodes in $In_i$, $V' = \bigcup_{i=1}^{n+1} In_i$ and is constructed using extraction Algorithm 6. An edge $e(u, u', a)$ is added to $E'$ whenever traversing BDD $B$ from $u$ wrt. encoding of $a$ ends in $u' \neq \mathbf{0}$. Traversals are executed using Algorithm 7. Starting from $u$, in each step the algorithm traverses BDD by taking the *low* branch when corresponding bit $a_i = 0$ or *high* branch when $a_i = 1$. Traversal takes at most $k_i$ steps, terminating as soon as it reaches a node labeled with a different CSP variable. The MDD extracted from a log-BDD in Figure 4(a) is shown in Figure 4(b).

---

**Algorithm 6**: Extract MDD.

**Data**: BDD $B(V, E)$
$E' \leftarrow \{\}, V' \leftarrow \{r\}$;
**foreach** $i = 1, \ldots, n$ **do**
    **foreach** $u \in In_i$ **do**
        **foreach** $a \in D_i$ **do**
            $u' \leftarrow Traverse(u, a)$;
**1**            **if** $u' \neq \mathbf{0}$ **then**
                $E' \leftarrow E' \cup \{(u, u', a)\}$;
                $V' \leftarrow V' \cup \{u'\}$
return $(V', E')$;

---

**Algorithm 7**: Traverse BDD.

**Data**: BDD $B(V, E)$, $u$, $a$
$i \leftarrow cvar(u)$;
$(a_1, \ldots, a_{k_i}) \leftarrow enc_i(v)$;
**repeat**
    $s \leftarrow pos(u)$;
    **if** $a_s = 0$ **then**
        $u \leftarrow low(u)$;
    **else**
        $u \leftarrow high(u)$;
**until** $cvar(u) \neq i$ ;
return $u$;

---

Since each traversal (in line 1 of Algorithm 6) takes $O(\lceil log|D_i| \rceil)$ steps, the running time for the MDD extraction is $O(\sum_{i=1}^{n} |In_i| \cdot |D_i| \cdot \lceil log|D_i| \rceil)$. The resulting MDD $M(V', E')$ has at most $O(\sum_{i=1}^{n} |In_i| \cdot |D_i|)$ edges because we add at most $|D_i|$ edges for every node $u \in In_i$. Since we keep only nodes in $In_i$, $|V'| = \sum_{i=1}^{n} |In_i| \leq |V|$.

## 4.3 Input Model and Implementation Details

An important factor for usability of our approach is the easiness of specifying the input CSP model. BDD packages are callable libraries with no default support for CSP-like input language. To the best of our knowledge, the only open-source BDD-compilation tool that accepts as an input a CSP-like model is *CLab* (Jensen, 2007). It is a configuration interface on top of a BDD package *BuDDy* (Lind-Nielsen, 2001). CLab constructs a BDD for each input constraint and conjoins them to get the final BDD. Furthermore CLab generates a BDD using log-encoding with clustered ordering which suits well our extraction approach. Therefore, our compilation approach is based on using CLab to specify the input model and generate a BDD that will be used by our extraction Algorithm 6.

Note that after extracting the MDD, we preprocess it for efficient online querying. We expand the long edges and merge isomorphic nodes to get a *merged* MDD. We then *translate* it into a more efficient form for online processing. We rename BDD node names to indexes from $0, \ldots, |V|$, where root has index $0$ and terminal **1** has index $|V|$. This allows for subsequent efficient implementation of $U$ and $D$ labels, as well as an efficient access to children and parent edges of each node. In our initial experiments we got an order of magnitude speed-up of `wCVD` queries after we switched from BDD node names (which required using less efficient mapping for $U$, $D$, $Ch$ and $P$ structures).

## 5. Interactive Configuration With Multiple Costs

In a number of domains, a user should configure in the presence of multiple cost functions which express often conflicting objectives that a user wants to achieve simultaneously. For example, when configuring a product, a user wants to minimize the price, while maximizing the quality, reducing the ecological impact, shortening delivery time etc. We assume therefore that in addition to the CSP model $(X, D, F)$ whose solution space is represented by a merged MDD $M$, we are given $k$ additive cost functions

$$c_i(x_1, \ldots, x_n) = \sum_{j=1}^{n} c_{ij}(x_i), \ i = 1 \ldots, k$$

expressing multiple objectives. Multi-cost scenarios are often considered within the *multi-criteria optimization* framework (Figueira et al., 2005; Ehrgott & Gandibleux, 2000). It is usually assumed that there is an optimal (but unknown) way to aggregate multiple objectives into a single objective function that would lead to a solution that achieves the best balance in satisfying various objectives. The algorithms sample few *efficient* solutions (nondominated wrt. objective criteria) and display them to the user. Through user input, the algorithms learn how to aggregate objectives more adequately which is then used for the next sampling of efficient solutions etc. In some approaches a user is asked to explicitly assign weights $w_i$ to objectives $c_i$ which are then aggregated through weighted summation $c = \sum_{i=1}^{k} w_i c_i$.

While adopting these techniques to run over a compiled representation of solution space would immediately improve their complexity guarantees and would be useful in many scenarios where multi-criteria techniques are traditionally used, we believe that in a configuration setting, a more explicit control over variable values is needed. A user should easily explore the effect of assigning various variable values on other variables as well as cost functions. We therefore suggest to directly extend our `wCVD` query so that a user could explore the effect of cost restrictions in the same way he explores interactions between regular variables. The key query that we want to deliver is computing valid domains wrt. multiple cost restrictions:

**Definition 5** (k-wCVD) *Given a CSP model $(X, D, F)$, additive cost functions $c_j : D \to \mathbb{R}$, and maximal costs $K_j$, $j = 1, \ldots, k$, for a given partial assignment $\rho$, compute:*

$$VD_i[\rho, \{K_j\}_{j=1}^k] = \{a \in D_i \mid \exists \rho'. (\rho' \models F \text{ and } \rho \cup \{(x_i, a)\} \subseteq \rho' \text{ and } \bigwedge_{j=1}^k c_j(\rho') \leq K_j)\}$$

We are particularly interested in two-cost configuration as it is more likely to occur in practice and has strong connections to existing research in solving Knapsack problems and multi-criteria optimization. In the reminder of the section we will first discuss the complexity of 2-wCVD queries and then develop a practical implementation approach. We will then discuss the general k-wCVD query.

## 5.1 Complexity of 2-wCVD query

We assume that as an input to the problem we have a merged MDD $M$, additive cost functions $c_1, c_2$ and cost bounds $K_1, K_2$. The first question is whether it is possible for some restricted forms of additive cost functions $c_1, c_2$ to implement 2-wCVD in polynomial time. For this purpose we formulate a decision-version of the 2-wCVD problem:

**Problem 1** (2-wCVD-SAT) *Given CSP $(X, D, F)$ and MDD $M$ representation of its solution space, and given two additive cost functions $c_i(x) = \sum_{j=1}^n c_{ij}(x_j)$, $i = 1, 2$ with cost restrictions $K_1, K_2$, decide whether $F \wedge c_1(x) \leq K_1 \wedge c_2(x) \leq K_2$ is satisfiable.*

Unfortunately, the answer is no even if both constraints involve only positive coefficients, and have binary domains. To show this we reduce from the well-known *Two-Partition Problem* (TPP) which is NP-hard (Garey & Johnson, 1979). For a given set of positive integers $S = \{s_1, \ldots, s_n\}$, the TPP asks to decide whether it is possible to split a set of indexes $I = \{1, \ldots, n\}$ into two sets $A$ and $I \setminus A$ such that the sum in each set is the same: $\sum_{i \in A} s_i = \sum_{i \in I \setminus A} s_i$.

**Proposition 3** *The 2-wCVD-SAT problem defined over Boolean variables and involving only linear cost functions with positive coefficients is NP-hard.*

**Proof 3** *We show the stated by reduction from TPP. In order to reduce TPP to two-cost configuration we introduce $2n$ binary variables $x_1, \ldots, x_{2n}$ such that $i \in A$ if and only if $x_{2i-1} = 1$ and $i \in A \setminus I$ if and only if $x_{2i} = 1$. We construct an MDD for $F = \{x_1 \neq x_2, \ldots, x_{2n-1} \neq x_{2n}\}$ and introduce two linear cost functions with positive coefficients, $c_1(x) = \sum_{i=1}^n s_i \cdot x_{2i-1}$ and $c_2(x) = \sum_{i=1}^n s_i \cdot x_{2i}$. The overall capacity constraints are set to $K_1 = K_2 = \sum_{i \in I} s_i/2$. By setting $A = \{i \in I \mid x_{2i-1} = 1\}$ it is easily seen that $F \wedge c_1(x) \leq K_1 \wedge c_2(x) \leq K_2$ is satisfiable if and only if the TPP has a feasible solution. Hence, if we were able to solve 2-wCVD-SAT with Boolean variables and positive linear cost functions in polynomial time, we would also be able to solve the TPP problem polynomially.*

## 5.2 Pseudo-Polynomial Scheme for 2-wCVD

In the previous subsection we demonstrated that answering 2-wCVD queries is NP-hard even for the simplest class of positive linear cost functions over Boolean domains. Hence, there is no hope of solving `2-wCVD` with guaranteed polynomial execution time unless $P = NP$. However, we still want to provide a practical solution to the `2-wCVD` problem. We hope to avoid worst-case performance by exploiting the specific nature of the cost-functions we are processing. In this subsection we therefore show that `2-wCVD` can be solved in pseudo-polynomial time by extending our labeling approach from Section 3.2. Furthermore, we show how to adopt advanced techniques used for the Knapsack problem (Kellerer, Pferschy, & Pisinger, 2004).

### 5.2.1 Overall Approach

Our algorithm runs analogous to the single-cost approach developed in Section 3.2. After restricting the MDD wrt. a current assignment, we calculate upstream and downstream costs $U, D$ (which are no longer constants but lists of tuples), and use them to check for each edge $e$, whether $v(e)$ is in a valid domain.

For a given edge $e : u \rightarrow u'$, labeled with costs $c_1(e), c_2(e)$, it follows $v(e) \in VD_i$ iff there are paths $p : r \rightsquigarrow u$, and $p' : u' \rightsquigarrow \mathbf{1}$ such that $c_1(p) + c_1(e) + c_1(p') \leq K_1$ and $c_2(p) + c_2(e) + c_2(p') \leq K_2$. At each node $u$ it suffices to store two sets of labels:

$$U[u] = \{(c_1(p), c_2(p)) \mid p : r \rightsquigarrow u\}$$

$$D[u] = \{(c_1(p), c_2(p)) \mid p : u \rightsquigarrow \mathbf{1}\}$$

Then, for given cost restrictions $K_1, K_2$, and an edge $e : u \rightarrow u', u \in V_i$, domain $VD_i[K_1, K_2]$ contains $v(e)$ if for some $(a_1, a_2) \in U[u]$ and $(b_1, b_2) \in D[u]$ it holds

$$a_1 + c_1(e) + b_1 \leq K_1 \wedge a_2 + c_2(e) + b_2 \leq K_2 \tag{9}$$

### 5.2.2 Exploiting Pareto Optimality

While in the single-cost case it was sufficient to store at $U[u], D[u]$ only the minimal value (the cost of the shortest path to root/terminal), in multi-cost case we need to store multiple tuples. The immediate extension would require storing at most $K_1 \cdot K_2$ tuples at each node. However, we need to store only *non-dominated tuples* in $U$ and $D$ lists. If there are two tuples $(a_1, a_2)$ and $(a_1', a_2')$ in the same list such that

$$a_1 \leq a_1' \text{ and } a_2 \leq a_2'$$

then we may delete $(a_1', a_2')$ as if test (9) succeeds for $(a_1', a_2')$ it will also succeed for $(a_1, a_2)$. The remaining entries are the costs of *pareto-optimal* solutions. A solution is pareto-optimal wrt. solution set $S$ and cost functions $c_1, c_2$ if it is not possible to find a cheaper solution in $S$ with respect to one cost without increasing the other. Path $p : r \rightsquigarrow \mathbf{1}$ represents a pareto-optimal solution in *Sol* iff for each node $u$ on the path, both sub-paths $p_1 : r \rightsquigarrow u$ and $p_2 : u \rightsquigarrow \mathbf{1}$ are pareto-optimal wrt. the sets of paths $\{p : r \rightsquigarrow u\}$ and $\{p : u \rightsquigarrow \mathbf{1}\}$ respectively. Hence, for each node $u$ it suffices to store:

$$U[u] = \{(c_1(p), c_2(p)) \mid p : r \rightsquigarrow u, \forall_{p' : r \rightsquigarrow u}(c_1(p) \leq c_1(p') \vee (c_2(p) \leq c_2(p'))\}$$

$$D[u] = \{(c_1(p), c_2(p)) \mid p : u \rightsquigarrow \mathbf{1}, \forall_{p' : u \rightsquigarrow \mathbf{1}}(c_1(p) \leq c_1(p') \vee (c_2(p) \leq c_2(p'))\}$$

Note that due to pareto-optimality, for each $a_1 \in \{0, \ldots, K_1\}$ and each $a_2 \in \{0, \ldots, K_2\}$ there can be at most one tuple in $U$ or $D$ where the first coordinate is $a_1$ or the second coordinate is $a_2$. Therefore, for each node $u$, $U[u]$ and $D[u]$ can have at most $\min\{K_1, K_2\}$ entries. Hence, the space requirements of our algorithmic scheme are in worst case $O(|V| \cdot K)$ where $K = \min\{K_1, K_2\}$.

### 5.2.3 Computing $U$ and $D$ Sets

We will now discuss how to compute the $U$ and $D$ sets efficiently by utilizing advanced techniques for solving Knapsack problems (Kellerer et al., 2004). We recursively update $U$ and $D$ sets in a layer by layer manner as shown in Algorithm 8. The critical component of each recursion step in the algorithm is *merging lists* in lines 2 and 4. In this operation a new list is formed such that all dominated tuples are detected and eliminated. In order to do this efficiently, it is critical to keep both $U$ and $D$ lists *sorted* wrt. the first coordinate, i.e.

$$(a_1, a_2) \prec (a_1', a_2') \equiv a_1 < a_2.$$

If $U$ and $D$ are sorted, they can be merged in $O(K)$ time using the list-merging algorithm for Knapsack optimization from (Kellerer et al., 2004, Section 3.4).

---

**Algorithm 8**: Update $U, D$ labels.

**Data**: MDD $M$, Cost functions $c_1, c_2$, Bounds $K_1, K_2$

$U[\cdot] = \{(\infty, \infty)\}$, $U[r] = \{(0, 0)\}$;

**foreach** $i = 1, \ldots, n$ **do**
    **foreach** $u \in V_i$ **do**
        **foreach** $e : u \to u'$ **do**
            $S \leftarrow \emptyset$;
            **foreach** $(a_1, a_2) \in U[u]$ **do**
                **if** $a_1 + c_1(e) \leq K_1 \wedge a_2 + c_2(e) \leq K_2$ **then**
**1**                    $S \leftarrow S \cup (a_1 + c_1(e), a_2 + c_2(e))$;
**2**            $U[u'] \leftarrow MergeLists(S, U[u'])$;

$D[\cdot] = \{(\infty, \infty)\}$, $D[\mathbf{1}] = \{(0, 0)\}$;

**foreach** $i = n, \ldots, 1$ **do**
    **foreach** $u \in V_i$ **do**
        **foreach** $e : u \to u'$ **do**
            $S \leftarrow \emptyset$;
            **foreach** $(a_1, a_2) \in D[u']$ **do**
                **if** $a_1 + c_1(e) \leq K_1 \wedge a_2 + c_2(e) \leq K_2$ **then**
**3**                    $S \leftarrow S \cup (a_1 + c_1(e), a_2 + c_2(e))$;
**4**            $D[u] \leftarrow MergeLists(S, D[u])$;

---

The time complexity is determined by populating list $S$ (in lines 1 and 3) and merging (in lines 2 and 4). Each of these updates takes $O(K)$ in worst case. Since we perform these updates for each edge $e \in E$, the total time complexity of Algorithm 8 is $O(|E| \cdot K)$ in the worst case.

### 5.2.4 Valid Domains Computation

Once the $U, D$ sets are updated we can extract valid domains in a straightforward manner using Algorithm 9. For each edge $e : u \rightarrow u'$ the algorithm evaluates whether $v(e) \in VD_i$ in worst case $O(|U[u]| \cdot |D[u']|) = O(K^2)$ steps. Hence, valid domain extraction takes in worst case $O(|E| \cdot K^2)$ steps.

---

**Algorithm 9**: Compute valid domains.

**Data**: MDD $M$, Cost functions $c_1, c_2$, Cost bounds $K_1, K_2$, Labels $U, D$

**foreach** $i = 1, \ldots, n$ **do**
  $VD_i \leftarrow \emptyset$;
  **foreach** $u \in V_i$ **do**
   **foreach** $e : u \rightarrow u'$ **do**
    **foreach** $(a_1, a_2) \in U[u], (b_1, b_2) \in D[u']$ **do**
     **if** $a_1 + c_1(e) + b_1 \leq K_1 \wedge a_2 + c_2(e) + b_2 \leq K_2$ **then**
      $VD_i \leftarrow VD_i \cup \{v(e)\}$;
      **break**;

---

However, we can improve the running time of valid domains computation by exploiting (1) pareto-optimality and (2) the fact that the sets $U, D$ are sorted. It is critical to observe that given an edge $e : u \rightarrow u'$, for each $(a_1, a_2) \in U[u]$ it suffices to perform the validity test (9) only for a tuple $(b_1^*, b_2^*) \in D[u']$, where $b_1^*$ is a maximal first coordinate satisfying $a_1 + c_1(e) + b_1 \leq K_1$, i.e.

$$b_1^* = max\{b_1 \mid (b_1, b_2) \in D[u'], \ a_1 + c_1(e) + b_1 \leq K_1\}.$$

Namely, if the test succeeds for some $(b_1', b_2')$ where $b_1' < b_1^*$, it will also succeed for $(b_1^*, b_2^*)$ since due to pareto-optimality, $b_1' < b_1^* \Rightarrow b_2^* < b_2'$ and hence $a_2 + c_2(e) + b_2^* < a_2 + c_2(e) + b_2' \leq K_2$. Since the lists are sorted, comparing all relevant tuples can be performed efficiently by traversing $U[u]$ in increasing order, while traversing $D[u']$ in decreasing order. Algorithm 10 implements the procedure.

---

**Algorithm 10**: Extract edge value.

**Data**: MDD $M$, Cost constraints $c_1, c_2$, Bounds $K_1, K_2$, Edge $e : u \rightarrow u'$ in $E_i$

$\mathbf{a}(a_1, a_2) = U[u].begin()$;
$\mathbf{b}(b_1, b_2) = D[u'].end()$;
**while** $\mathbf{a} \neq \top \wedge \mathbf{b} \neq \perp$ **do**
  **if** $a_1 + c_1(e) + b_1 > K_1$ **then**
   $\mathbf{b}(b_1, b_2) \leftarrow D[u'].previous()$;
   **continue**;
1   **else if** $a_1 + c_1(e) + b_1 \leq K_1 \wedge a_2 + c_2(e) + b_2 \leq K_2$ **then**
   $VD_i \leftarrow VD_i \cup \{v(e)\}$;
2    **return**;
  $\mathbf{a}(a_1, a_2) \leftarrow U[u].next()$;

---

The algorithm relies on several list operations. Given list $L$ of sorted tuples, operations $L.begin()$ and $L.end()$ return the first and the last tuple respectively wrt. the list ordering.

Operations $L.next()$ and $L.previous()$ return the next and the previous element in the list wrt. the ordering. Elements $\top$ and $\bot$ indicate two special elements that appear after the last and before the first element in the list respectively. They indicate that we have passed beyond the boundary of the list. The algorithm terminates (line 2) as soon as the test succeeds. Otherwise, it keeps iterating over tuples until we have processed either the last tuple in $U[u]$ or the first tuple in $D[u']$. In that case the algorithm terminates as it is guaranteed that $v(e) \notin VD_i$. In each step, we traverse at least one element from $U[u]$ or $D[u']$. Hence, in total we can execute at most $U[u] + D[u'] \leq 2K$ operations. Therefore, the time complexity of single edge traversal is $O(K)$ and the complexity of valid domains computation of Algorithm 9 (after replacing the quadratic loop with Algorithm 10) is $O(|E| \cdot K)$ where $K = \min\{K_1, K_2\}$.

In conclusion, we have developed a pseudo-polynomial scheme for computing valid domains wrt. two cost functions (2-wCVD). The space complexity is dominated by storing $U$ and $D$ sets at each node. In worst case we have to store $O(|V| \cdot K)$ entries. The time complexity to compute $U$ and $D$ labels and extract valid domains takes $O(|E| \cdot K)$ steps. The overall interaction is similar to the single-cost approach. After assigning a variable, we have to recompute the labels as well as extract domains. If we tighten cost restrictions $K_1, K_2$ to $K_1' \leq K_1, K_2' \leq K_2$ we only need to extract domains. However, if we relax either of the cost restrictions, such as $K_1' > K_1$ we need to recompute the labels as well. More precisely, labels $U, D$ need to be recomputed only if $K_1 > K_1^{max}$ where $K_1^{max}$ was the initial cost restriction after the last assignment.

### 5.2.5 FURTHER EXTENSIONS

Note that our approach can, in principle, be extended to handle general k-wCVD query for a fixed $k$. Lists $U$ and $D$ would contain the set of non-dominated $k$-tuples, ordered such that: $(a_1, \ldots, a_k) \prec (a_1', \ldots, a_k')$ iff for the smallest coordinate $j$ for which $a_j \neq a_j'$ it holds $a_j < a_j'$. Both the list merging as well as valid domains extraction would be directly generalized to operate over such ordered sets, although the time complexity for testing dominans will increase. The worst-case complexity would depend on the size of an efficient frontier, which for $k$ cost functions with cost bounds $K$ is bounded by $O(K^{k-1})$. In practice however, we could expect that the number of non-dominated tuples be much smaller, especially for cost functions over smaller scopes and with smaller coefficients. Note that our approach can also be extended to accommodate non-additive cost functions by expanding the MDD to accommodate non-unary labels in the same fashion as discussed in Section 3.4.

## 5.3 Approximation Scheme for 2-wCVD

In this subsection we analyze the complexity of answering 2-wCVD queries in approximative manner, i.e. how can we improve running time guarantees by settling for an approximate solution. Assume that one of the constraints $K_2$ is fixed while the second constraint may be exceeded with a small tolerance $(1+\epsilon)K_1$. For example, a user might be willing to tolerate a small increase in price as long as strict quality restrictions are met. In this section we present a fully polynomial time approximation scheme (FPTAS) for calculating valid domains in time $O(En\frac{1}{\epsilon})$ for this problem. The FPTAS should satisfy that no feasible solution with respect to the original costs should be fathomed, and that any feasible configuration found by use of the FPTAS in the domain restriction should satisfy the cost constraint within $(1 + \epsilon)K_1$. Finally, the FPTAS should have running time polynomial in $1/\epsilon$ and the input size.

In order to develop the FPTAS we use a standard scaling technique (Schuurman & Woeginger, 2005) originally presented by Ibarra and Kim (1975). Given an $\epsilon$, let $n$ be the number of decision variables. Set $T = \epsilon K_1/(n + 1)$ and determine new costs $\tilde{c}_1(e) = \lfloor c_1(e)/T \rfloor$ and new bounds $\tilde{K}_1 = \lceil K_1/T \rceil$. We then perform the valid domains computation (label updating and domain extraction) as described in Section 5.2, using the scaled weights. The following propositions prove that we obtained a FPTAS scheme.

**Proposition 4** *The running time of valid domains computation is $O(\frac{1}{\epsilon}En)$*

**Proof 4** *We may assume that $\tilde{K}_1 < K_2$ as otherwise we may interchange the two costs. The running time becomes*

$$O(E\tilde{K}_1) = O(EK_1/T) = O(EK_1 \frac{n+1}{\epsilon K_1}) = O(En\frac{1}{\epsilon})$$

*since $n \leq V$ this is polynomial in the input size $O(V + E)$ and the precision $\frac{1}{\epsilon}$.*

**Proposition 5** *If a solution was feasible with respect to the original costs, then it is also feasible with respect to the scaled costs.*

**Proof 5** *Assume that $\sum_{e \in p} c_1(e) \leq K_1$. Then*

$$\sum_{e \in p} \tilde{c}_1(e) = \sum_{e \in p} \lfloor c_1(e)/T \rfloor \leq \frac{1}{T} \sum_{e \in p} c_1(e) \leq \frac{1}{T}K_1 \leq \lceil \frac{1}{T}K_1 \rceil = \tilde{K}_1$$

**Proposition 6** *Any solution that was feasible with respect to the scaled costs $\tilde{c}_1(e)$ satisfies original constraints within $(1 + \epsilon)K_1$.*

**Proof 6** *Assume that $\sum_{e \in p} \tilde{c}_1(e) \leq \tilde{C}_1$. Then*

$$\begin{aligned}
\sum_{e \in p} c_1(e) &= T \sum_{e \in p} c_1(e)/T \leq T \sum_{e \in p}(\lfloor c_1(e)/T \rfloor + 1) \leq T \sum_{e \in p} \tilde{c}_1(e) + Tn \\
&\leq T\tilde{K}_1 + Tn = T\lceil K_1/T \rceil + Tn \leq T(K_1/T + 1) + Tn \\
&= K_1 + T(n + 1)
\end{aligned}$$

*Since $T = \epsilon K_1/(n + 1)$ we get*

$$\sum_{e \in p} c_1(e) \leq K_1 + (n + 1)\epsilon K_1/(n + 1) = (1 + \epsilon)K_1$$

*which shows the stated.*

The time complexity can be further improved using techniques from Kellerer et al. (2004) for the Knapsack Problem, but we are here only interested in showing the existence of a FPTAS.

By the considerations in previous subsections we have fully analyzed the complexity of answering `2-wCVD` queries. We first showed that this is an NP-hard problem. We then developed a pseudo-polynomial scheme for solving it, and finally we devised a fully polynomial time approximation scheme. Even though we cannot provide polynomial running-time guarantees, based on these considerations, we can hope to provide a reasonable performance for practical instances, as it will be demonstrated in Section 6.

## 5.4 Complexity of `k-wCVD` Query

We conclude this section by discussing complexity of general `k-wCVD` queries. While our practical implementation efforts are focused on implementing `2-wCVD` queries, or other `wCVD` queries where the number of cost constraints is known in advance, for completeness we consider a generic problem of delivering `k-wCVD` for *arbitrary* $k$, i.e. where $k$ is part of the input to the problem.

We will prove now that for such a problem there is no pseudo-polynomial scheme unless NP=P. We will show that decision version of such problem `k-wCVD-SAT` is *NP-hard in the strong sense* (Garey & Johnson, 1979) by reduction from the *bin-packing problem* (BPP) which is strongly NP-hard (Garey & Johnson, 1979). In the decision form the BPP asks whether a given set of numbers $s_1, \ldots, s_n$ can be placed into $k$ bins of size $K$ each. Notice, that we cannot use reduction below for showing NP-hardness of `2-wCVD-SAT`, since $k$ is a part of the input in BPP.

**Theorem 7** *The `k-wCVD-SAT` problem with variable $k$, is strongly NP-hard.*

**Proof 7** *For a given instance of BPP we reduce it to a `k-wCVD-SAT` instance as follows: We construct an MDD for a CSP$(X, D, F)$ over $n$ variables $X = \{x_1, \ldots, x_n\}$ each with a domain of size $k$, $D_i = \{1, \ldots, k\}$, $i = 1, \ldots, n$. We set $F = \emptyset$, so that resulting MDD allows all assignments. It has $n$ nonterminal nodes $u_1, \ldots, u_n$ corresponding to the numbers $s_1, \ldots, s_n$. Between two nodes $u_i, u_{i+1}$ we have $k$ edges with costs $(c_1(e), c_2(e), \ldots, c_k(e))$ set to*

$$(s_i, 0, \ldots, 0), (0, s_i, 0, \ldots, 0), (0, 0, s_i, 0, \ldots, 0), \ldots, (0, \ldots, s_i),$$

*The first node $u_1$ is the root $u_1 = r$ while the last node $u_n$ is connected to the terminal $u_{n+1} = \mathbf{1}$. The overall capacity constraints are $(K_1, \ldots, K_k) = (K, \ldots, K)$.*

*It is easily seen that we may find a path from $r$ to $\mathbf{1}$ if and only if the BPP has a feasible solution. Since the BPP is strongly NP-hard we have shown that `k-wCVD-SAT` also is strongly NP-hard.*

## 6. Experimental Evaluation

We implemented our compilation scheme and the algorithms for `wCVD` and `2-wCVD` queries. We performed a number of experiments to evaluate the applicability of our approach as well as to confirm various hypotheses made throughout the paper. We used two sets of instances whose properties are presented in Table 1. The first set corresponds to real-world configuration problems available at configuration benchmarks library CLib[2]. These are CSP models with configuration constraints. They correspond to highly structured configuration problems with a huge number of similar solutions. The second set of instances represents product-catalogue datasets used by Nicholson, Bridge, and Wilson (2006). These catalogues are defined explicitly, as tables of solutions. They represent a much smaller and sparser set of solutions.

---

2. `http://www.itu.dk/research/cla/externals/clib/`

| Instance | Sol | X | $d_{min}$ | $d_{max}$ | $d_{avg}$ |
|---|---|---|---|---|---|
| ESVS | $2^{31}$ | 26 | 2 | 61 | 5 |
| FS | $2^{24}$ | 23 | 2 | 51 | 5 |
| Bike2 | $2^{26}$ | 34 | 2 | 38 | 6 |
| PC2 | $2^{20}$ | 41 | 4 | 34 | 9 |
| PC | $2^{20}$ | 45 | 2 | 33 | 8 |
| Big-PC | $2^{83}$ | 124 | 2 | 66 | 12 |
| Renault | $2^{41}$ | 99 | 2 | 42 | 4 |
| Travel | 1461 | 7 | 4 | 839 | 134 |
| Laptops | 683 | 14 | 2 | 438 | 42 |
| Cameras | 210 | 9 | 5 | 165 | 40 |
| Lettings | 751 | 6 | 2 | 174 | 45 |

Table 1: First seven instances are real-world configuration problems available at configuration benchmarks library CLib. Remaining four instances are product catalogues used by Nicholson et al. (2006). For each instance we provide the number of solutions *Sol*, number of variables *X*, the minimal, maximal and average domain size.

## 6.1 MDD Size

In the first set of experiments, for each instance we generated a log-encoded BDD $B$ using CLab (Jensen, 2007). We then *extracted* a corresponding MDD $M$ from $B$. Finally, we expanded long edges in $M$ and merged isomorphic nodes to generate a merged MDD $M'$. We compare the sizes of $B$, $M$ and $M'$ in Table 2. For each structure we provide the number of nodes $V$ and edges $E$. We also provide the size of the BDD $B$. We conclude from the table that both BDDs and MDDs are exponentially smaller than the size of the solution space for configuration instances while not as significantly smaller for more diverse product configuration catalogues. Furthermore, we can see that the number of edges in merged MDDs $M'$ is not significantly larger in comparison to extracted MDDs $M$. Hence, due to simpler online algorithms, using merged MDDs seems well suited for online reasoning. We can also see that multi-valued encoding in many cases reduces the number of nodes and edges in comparison to BDDs. Even though compilation times are less important since the generation of the MDD is performed offline, it is worth noting that for the largest instance, Renault, it took around 2min and 30sec to compile the instance into a BDD and extract an MDD.

### 6.1.1 Encoding Cost Explicitly

We also investigated the impact of encoding cost information explicitly into an MDD. For each instance we compared the size of the MDD without and with cost variables ($M$ and $M^c$ respectively). For configuration benchmarks we introduce an additional variable $y \in [0, 10000]$ such that $y = \sum_{i=1}^n a_i x_i$ where coefficients $a_i$ are randomly drawn from the interval $[0, 50]$. We put variable $y$ as the last in the ordering since for other positions we get MDDs of similar size, and putting $y$ at the end allows easier theoretical analysis. Since product catalogues already contain the cost variable $y$ (price), we produce a cost-oblivious version $M$ by existentially quantifying $y$, $M = \exists_y M^c$.

| Instance | $V_B$ | $E_B$ | KB | $V_M$ | $E_M$ | $V_{M'}$ | $E_{M'}$ |
|----------|-------|-------|-----|-------|-------|----------|----------|
| ESVS | 306 | 612 | 5 | 87 | 202 | 96 | 220 |
| FS | 3,044 | 6,088 | 41 | 753 | 1,989 | 767 | 2017 |
| Bike2 | 3,129 | 6,258 | 56 | 853 | 1,726 | 933 | 1886 |
| PC2 | 13,332 | 26,664 | 237 | 3,907 | 6,136 | 3907 | 6136 |
| PC | 16,494 | 32,988 | 298 | 4875 | 7989 | 4875 | 7989 |
| Big-PC | 356,696 | 713,392 | 7,945 | 100,193 | 132,595 | 100,272 | 132,889 |
| Renault | 455,796 | 911,592 | 9,891 | 283,033 | 334,008 | 329,135 | 426,212 |
| Travel | 8479 | 16,958 | 154 | 1469 | 2928 | 1469 | 2928 |
| Laptops | 9528 | 19,056 | 172 | 2033 | 2713 | 2033 | 2713 |
| Cameras | 4274 | 8,548 | 71 | 791 | 999 | 791 | 999 |
| Lettings | 2122 | 4,244 | 36 | 351 | 1099 | 351 | 1099 |

Table 2: Comparison between BDDs and MDDs for instances from Table 1. The second, third and fourth column give the number of non-terminal BDD nodes $V_B$, the number of edges $E_B$ and the size on disk of the BDD in kilobytes $KB$. The fifth and the sixth column give the number of vertices $V_M$ and edges $E_M$ in an MDD $M$ *extracted* from the BDD using Algorithm 6 on page 116. The final two columns provide the number of nodes and edges in a *merged MDD* ($M'$) where all long edges from extracted MDD $M$ have been expanded.

In Table 3 we compare the MDDs $M$ and $M^c$. For both structures we provide the number of edges as well as the representation size in kilobytes. We also show the size of cost range $C(Sol)$. We observe that for configuration instances that have a high level of sharing and compression, introducing cost information explicitly induces an order of magnitude increase in size even when the cost range $C(Sol)$ is limited (400 times increase for Bike2 instance). MDDs for the two largest instances could not be generated. However, for product catalogues which have much less sharing, removing cost information does not have a dramatic effect. In the worst case, the number of edges in $M^c$ is two times larger than in $M$. Hence, the experimental results confirm that introducing cost explicitly could have a dramatic effect for MDD representations of highly compressed solution spaces, usually implicitly defined through conjunction of combinatorial constraints. However, the effect of adding explicit cost information might be modest when the solution space is defined explicitly, as a (sparse) list of database entries, such as the case for product catalogues. Furthermore, the size of the cost range $C(Sol)$ needs not be significant for a large increase in size to take place.

## 6.2 Response Times for `wCVD` Queries

In the second set of experiments, we evaluated the performance of `wCVD` queries over merged MDD representations of configuration instances. We report the running times for both computing $U$ and $D$ labels (Algorithm 2) as well as computing valid domains (Algorithm 3). In Table 4 we report both average and worst-case running times over initial merged MDDs from Table 2. We also report the time necessary to restrict the MDD wrt. an assignment (Algorithm 1). We randomly create an additive cost function $c$ by assigning for each variable $x_i$ and each value $a \in D_i$ a cost $c_i(a)$ from $[0, 50]$. Valid domains are computed wrt. the maximal cost restriction $K$ that is set to a value larger than the the length of the longest MDD path wrt. cost function $c$. This ensures the longest execution time of Algorithm 3. Each data-point in the table is an average or a maximum over 1000 executions on a Fedora 9 operating system, using dual Quad core Intel Xeon processor running at 2.66 GHz. Only one core is used for each instance. Empirical evaluation demonstrates that response times are easily within acceptable interaction bounds even for the largest instances, where in worst case the MDD nodes are labeled within 0.13 seconds, valid domains are computed within 0.07 seconds and MDD is restricted wrt. an assignment within 0.28 seconds.

| Instance | E | KB | $E^c$ | KB | C(Sol) |
|----------|------:|------:|----------:|------:|------:|
| ESVS | 202 | 5 | 129,514 | 4,408 | 1,966 |
| FS | 1,989 | 41 | 407,662 | 12,767 | 1,497 |
| Bike2 | 1,726 | 56 | 693,824 | 31,467 | 3,008 |
| PC2 | 6,136 | 237 | 1,099,842 | 57,909 | 2,000 |
| PC | 7,989 | 298 | 1,479,306 | 70,900 | 2,072 |
| Big-PC | 132,595 | 7,945 | - | - | - |
| Renault | 334,008 | 9,891 | - | - | - |
| Travel | 1640 | 45 | 2928 | 154 | 839 |
| Laptops | 1592 | 80 | 2713 | 172 | 438 |
| Cameras | 725 | 44 | 999 | 71 | 165 |
| Lettings | 496 | 9 | 1099 | 36 | 174 |

Table 3: Effect of explicitly encoding cost information. The second and third column indicate the number of edges and the representation size in kilobytes for cost-oblivious MDD, while the fourth and fifth column show the same for the MDD containing cost information. Column $C(Sol)$ indicates the range of available costs over all solutions.

## 6.3 Response Times for `2-wCVD` Query

We generated analogous statistics for `2-wCVD` in Table 5. We tested the performance of our algorithms under the computationally most demanding circumstances: we operate over the original (fully-sized) MDD, even though during interaction it would be reduced due to user assignments. Furthermore, both cost functions $c_1, c_2$ have a global scope, and we use no cost restrictions when computing $U$ and $D$ labels (i.e. we ignore the condition in line 1 of Algorithm 10 and hence, $U[\mathbf{1}]$ and $D[r]$ correspond to an entire efficient frontier). Normally, cost functions would involve only a subset of variables and only a fraction of the labels on the efficient frontier (within restrictions $K_1, K_2$) would be relevant for the user. We generate cost functions $c_1, c_2$ by drawing costs $c_i(a)$ randomly from $[0, 50]$. For computing valid domains, we use restrictions $K_1, K_2$ larger than the lengths of corresponding longest paths, so that all possible solutions in the efficient frontier are allowed. This would lead to the longest execution time of Algorithm 9.

| | Labeling $U, D$ | | Valid domain | | Restrict | |
|---|---|---|---|---|---|---|
| **Instance** | **avg** | **max** | **avg** | **max** | **avg** | **max** |
| ESVS | 0.0001 | 0.01 | 0.0001 | 0.01 | 0.0001 | 0.01 |
| FS | 0.0001 | 0.01 | 0.0001 | 0.01 | 0.0002 | 0.01 |
| Bike2 | 0.0002 | 0.01 | 0.0001 | 0.01 | 0.0010 | 0.01 |
| PC2 | 0.0002 | 0.01 | 0.0002 | 0.01 | 0.0010 | 0.02 |
| PC | 0.0003 | 0.01 | 0.0003 | 0.01 | 0.0010 | 0.01 |
| Big-PC | 0.0210 | 0.04 | 0.0110 | 0.03 | 0.0400 | 0.08 |
| Renault | 0.0590 | 0.13 | 0.0310 | 0.07 | 0.1600 | 0.28 |

Table 4: Interaction time in seconds for `wCVD` queries. We report time required for computing $U$ and $D$ labels, valid domain computation and restriction wrt. a single assignment.

Our algorithms can easily handle the first five instances. For the largest two instances, if $U$ and $D$ labels are known, calculating valid domains can be done within a fraction of a second. Hence, a user can efficiently explore the effect of various cost restrictions $K_1, K_2$ wrt. a fixed partial assignment. After a user assigns a variable, recomputing $U$ and $D$ labels takes in total on average less than 0.85 seconds, or in worst case less than 1.4 seconds. While this is already within acceptable interaction times, the usability of the system can be further enhanced, e.g. by using a *layered display* of information: always reacting with the information that is fastest to compute (such as `CVD` or `wCVD`), and while the user is analyzing it, execute more time consuming operations. In particular, the entire efficient frontier is known as soon as $U$ labels are generated — in worst case within 0.64 seconds. At this stage, a user can explore the "cost-space" while $D$ labels are computed (on average within the next 0.79 seconds). Note that the running times can be reduced through a number of additional schemes, e.g. by computing $U$ and $D$ labels in parallel, if two or more processors are present.

| | Labeling $U$ | | Labeling $D$ | | Valid domain | |
|---|---|---|---|---|---|---|
| **Instance** | **avg** | **max** | **avg** | **max** | **avg** | **max** |
| ESVS | 0.0001 | 0.01 | 0.0002 | 0.01 | 0.0001 | 0.01 |
| FS | 0.0010 | 0.01 | 0.0020 | 0.02 | 0.0001 | 0.01 |
| Bike2 | 0.0010 | 0.02 | 0.0020 | 0.01 | 0.0001 | 0.01 |
| PC2 | 0.0030 | 0.02 | 0.0030 | 0.02 | 0.0005 | 0.01 |
| PC | 0.0050 | 0.02 | 0.0040 | 0.02 | 0.0008 | 0.01 |
| Big-PC | 0.2070 | 0.45 | 0.3160 | 0.60 | 0.0300 | 0.04 |
| Renault | 0.3470 | 0.64 | 0.4700 | 0.79 | 0.0700 | 0.08 |

Table 5: Interaction time in seconds for `2-wCVD` query.

Our empirical evaluation demonstrates the practical value of our approach. Even the NP-hard `2-wCVD` query can be implemented with response times suitable for interactive use, when applied to huge configuration instances. Note, however, that in order to achieve such performance it is critical to optimize MDD implementation as well as to utilize advanced list operation techniques. Our initial implementation efforts that failed to do so, led to response times measured in tens of seconds for the largest instances.

## 7. Related Work

There are several directions of related work. There is a large variety of representations investigated in the area of knowledge compilation that might be suitable for supporting interactive decision making with cost restrictions. There are also a number of approaches to handle multiple cost functions in multi-criteria optimization.

### 7.1 Compiled Knowledge Representation Forms

In this paper we used binary decision diagrams (BDDs) and multi-valued decision diagrams (MDDs) as compiled representations of our CSP model. However, there might be other compiled representations that might be more suitable for supporting interactive configuration. Any compiled representation that supports efficient *consistency* checking and *conditioning* would in theory support polytime interactive *configuration*. To calculate valid domains it suffices for each value to restrict the representation and check if it is consistent. Any representation that supports efficient *optimization* and conditioning would support polytime *cost restrictions*. It would suffice to restrict the representation with a value and check if the minimum is smaller than a threshold value. We will therefore briefly survey some of the related compiled representations and evaluate their suitability for our framework.

**Knowledge-Compilation Structures**. Probably the most well known framework for comparing various compiled forms of propositional theories is based on viewing them as special classes of *negation normal form* (NNF) languages (Darwiche & Marquis, 2002). NNFs are directed acyclic graphs where internal nodes are associated with conjunctions ($\wedge$) or disjunctions ($\vee$), while leaf nodes are labeled with literals ($x, \neg x$) or constants *true* or *false*. By imposing various restrictions we get subclasses of NNF languages that support efficient execution of various queries and transformations. More restrictive representations are *less succinct* i.e. they can be exponentially larger for some instances, but they support a larger number of queries and transformations in polytime. A comprehensive overview of such representations is presented by Darwiche and Marquis (2002).

The critical restriction that makes NNF languages more tractable is *decomposability*. It exploits variable independencies by enforcing that children of an $\wedge$-node have non-overlapping variable scopes. Hence, for a propositional formula $F = F_1 \wedge F_2$ such that $var(F_1) \bigcap var(F_2) = \emptyset$, to evaluate satisfiability of $F$ it suffices to independently evaluate $F_1$ and $F_2$. A resulting language is *decomposable negation normal form* (DNNF) which already supports in polytime two operations critical for calculating valid domains: *consistency* checking and *conditioning*. However, no general DNNF compiler exists. Current compilation approach based on exhaustive DPLL search with caching isomorphic nodes (Huang & Darwiche, 2005) constructs subsets of DNNF that satisfy an additional property of *determinism*. Any two children of an ∨-node are mutually exclusive. The resulting structure is called *deterministic decomposable negation normal form* (d-DNNF). This structure would be an interesting target for cost-configuration. For Boolean CSP models, additive cost functions could be efficiently optimized over d-DNNFs. For multi-valued models however, more research is necessary on how to encode finite-domain values in a way that allows efficient cost processing. The tool support for compiling d-DNNFs so far takes as an input only CNF formulas, and we are unaware of extensions allowing direct compilation of general CSP models.

Other known knowledge representation forms can be retrieved by enforcing additional properties. For example, by further enforcing that all nodes are *decision* nodes and that each variable is encountered at most once on each path (*read-once* property) we get *free BDDs* (FBDDs). After enforcing that all decision nodes appear wrt. fixed *ordering* we get *ordered BDDs* (OBDDs). In fact, the d-DNNF compiler of Huang and Darwiche (2005) can be specialized to compile OBDDs, which proved to be a valuable alternative way to BDD compilation.

**Weighted and Multi-Valued Knowledge Compilation Structures**. Most of the compiled representations for propositional theories have *valued* counterparts. Many of them can be seen as special cases of *valued NNFs* (VNNF) (Fargier & Marquis, 2007). Roughly, every valued counterpart is obtained by changing the semantics of nodes, from logical operators (such as ∧, ∨) to general operators ⊗ (that could be arithmetic, such as + and *). Values of functions represented by these structures are no longer in $\{0, 1\}$ but in $\mathbb{R}$. Furthermore, functions need not be defined over Boolean domains, but could take finite-domain values. In general, subsets of VNNF that satisfy decomposability and operator *distributivity* support efficient optimization (Fargier & Marquis, 2007) and could, in principle, be used to support cost configuration.

Construction of MDDs based on encoding into BDDs has been discussed by Srinivasan, Kam, Malik, and Brayton (1990). Amilhastre et al. (2002) augmented automata of Vempaty (1992) with edge weights to reason about optimal restorations and explanations. These weighted extensions correspond closely to our weighted MDDs since the variant of automata used by Vempaty (1992) is equivalent to merged MDDs (Hadzic et al., 2008). However, the weights are used to compute different queries and while we generate MDDs based on widely available BDD-packages, Vempaty (1992) does not report compilation tools used. *Semiring labeled decision diagrams* (SLDDs) (Wilson, 2005) label edges of an (unordered) MDD with values from a semiring and allow computation of a number of queries relevant for reasoning under uncertainty. Due to relaxed ordering, SLDDs are more succinct than our weighted MDDs and are therefore an attractive target for cost-based configuration. However, the proposal for now seems to be theoretic, and does not seem to be implemented. *Arithmetic circuits* are directed acyclic graphs where internal nodes are labeled with summation and multiplication operators while leaf nodes are labeled with constants or variables (Darwiche, 2002). They could be seen as a valued extension of d-DNNFs and hence are more succinct than SLDDs. Furthermore, they support efficient optimization when all coefficients are positive (in Bayesian context - they support efficient computing of *most probable explanations*). Compilation technology for ACs is not directly applicable to general CSP models, as it is used primarily for representing Bayesian networks. It is based on compiling d-DNNFs or tree clustering approaches (Darwiche, 2002, 2003). In our context, ACs might be use-ful when optimizing non-additive objective functions with multiplicative coefficients such as multi-linear functions induced by Bayesian networks. However, for purely propositional constraints over which an additive cost function should be optimized, a purely propositional representation form (such as d-DNNF) would be more adequate. Furthermore, efficient optimization queries based on ACs implicitly assume that all constants (at leaf nodes) are positive, which is the case when modeling Bayesian networks, but does not hold for general cost functions.

**Global Structure Approaches**. A number of techniques based on *tree-clustering* (Dechter & Pearl, 1989) and *variable-elimination* (Dechter, 1999) exploit variable independencies that are present globally in a CSP model. Both time and space complexity of these techniques turn out to be bounded exponentially in the size of an important graph-connectivity notion of *tree-width* (Bodlaender, 1993). While most of these techniques are geared towards enhancing search for a single (optimal) solution (adaptive consistency, bucket elimination etc), the same concepts can be utilized for compiling representations of all solutions. AND/OR MDDs (Mateescu et al., 2008) when restricted to Boolean variables are a subset of d-DNNF formulas, where variable labeling respects a *pseudo-tree* obtained by a *variable elimination* order. Due to utilization of variable independencies through $\wedge$-nodes, they are more succinct than MDDs and are therefore an attractive compilation target for cost-configuration. Furthermore, they are already extended to handle *weighted graphical models* to support Bayesian reasoning. However, publicly available tool support is limited and does not allow processing weighted CVD queries. *Tree-driven-automata* (Fargier & Vilarem, 2004) utilize tree clustering (Dechter & Pearl, 1989), to generate a partial variable ordering that is used to generate an automaton. Tree-driven-automata are equivalent to AND/OR MDDs and when restricted to the Boolean case they represent a subset of d-DNNF languages called *strongly ordered decomposable decision graphs* (SO-DDG) (Fargier & Marquis, 2006). Like AND/OR MDDs they are more succinct than MDDs and therefore are an interesting target for cost-configuration. However, tools for compiling tree-driven-automata are yet to become publicly available, and so far they have not been extended to handle costs. *Weighted cluster trees* of Pargamin (2003) are a weighted extension of cluster trees used to support interactive configuration with preferences. However, there is no publicly available compilation tool (an internal company-based implementation was presented), and the clusters are represented explicitly without utilizing compressions based on *local structure* through decision diagrams or other compiled representations. *Tree-of-BDDs* (ToB) (Subbarayan, 2008) directly exploit tree clustering by representing each cluster as a BDD. However, they do not support conditioning in polytime which is a fundamental transformation in supporting user interaction (assigning variables). However, they can be compiled for instances for which d-DNNF compilation fails, and empirical evaluation shows that on average conditioning times are short.

**BDD Extensions**. There is a large variety of *weighted* extensions of binary decision diagrams, that represent real-valued functions $f : \{0, 1\}^n \to \mathbb{R}$ rather than Boolean functions $f : \{0, 1\}^n \to \{0, 1\}$. These extensions are limited to Boolean variables and their adoption in future would have to consider encoding techniques of multi-valued variables that avoid explosion in size and support cost processing. Comprehensive overviews of these extensions are presented by Drechsler (2001), Wegener (2000), and Meinel and Theobald (1998). An immediate extension is in the form of *algebraic decision diagrams* (ADDs) (Bahar, Frohm, Gaona,

Hachtel, Macii, Pardo, & Somenzi, 1993), also known as *multi-terminal BDDs* (MTBDDs), that are essentially BDDs with multiple terminal nodes - one for each cost value. This is a structure-oblivious approach to encoding cost, much as our approach of explicitly encoding cost as a variable. The size grows quickly with increase of the number of terminals. Therefore a number of BDD extensions are introduced based on *labeling edges with weights*. They differ mostly on cost operators and decomposition types associated with nodes. *Edge-valued BDDs* (EVBDDs) (Lai & Sastry, 1992) label every edge with an additive cost value $c(e)$ so that for an edge $e : u \rightarrow u'$, the value $val(u) = c(e) + val(u')$ when $v(e) = 1$ (otherwise $val(u) = val(u')$). *Factored EVBDDs* (FEVBDDs) (Tafertshofer & Pedram, 1997) introduce multiplicative weights, so that when $v(e) = 1$, value $val(u) = c(e) + w(e) \cdot val(u')$ (otherwise $val(u) = val(u')$). *Affine ADDs* (AADDs) of Sanner and McAllester (2005) further introduce additive and multiplicative edge weights for any edge (regardless of $v(e)$). Then $val(u) = c(e) + w(e) \cdot val(u')$ for every edge. It has been shown that AADDs are a special case of valued NNFs (Fargier & Marquis, 2007).

An orthogonal extension of BDDs is to change decomposition type of nodes. OBDDs are based on *Shannon decomposition* $f_u = x_i f_{u_0} \vee \neg x_i f_{u_1}$. We can change this decomposition type to *positive Davio* (pD) decomposition $f_u = f_0 \oplus x_i f_1$ or *negative Davio*(nD) decomposition $f_u = f_0 \oplus \neg x_i f_1$. By using pD decomposition we get *ordered functional decision diagrams* (OFDDs) (Kebschull & Rosenstiel, 1993). These structures are incomparable to OBDDs, i.e. they might be exponentially larger or smaller than OBDDs depending on the instance. However, *ordered Kronecker functional decision diagrams* (OKFDDs)(Drechsler, Sarabi, Theobald, Becker, & Perkowski, 1994) allow all three decomposition types, thus generalizing both OBDDs and OFDDs. Extending OFDDs with additive edge weights leads to *binary moment diagrams* (BMDs) (Bryant & Chen, 1995), adding also multiplicative edge weights leads to *multiplicative binary moment diagrams* (\*BMDs). Analogously, by extending OKFDDs with additive and multiplicative edge weights we get *Kronecker binary moment diagrams* (KBMDs) and $K^*BMDs$ respectively (Drechsler, Becker, & Ruppertz, 1996).

It is unclear whether Boolean structures with advanced cost labeling schemes can be used directly to represent multi-valued CSP models with cost functions. However, we could compare the generalizations of such labeling schemes to multi-valued structures. A multi-valued generalization of EVBDDs would correspond roughly to our weighted MDDs. However, introducing both additive and multiplicative weights as in AADDs would correspond to a generalization of our labeling scheme that could prove to be useful for labeling multi-linear cost functions. Namely, through introduction of multiplicative weights there would be more subgraph sharing, and not as many nodes would have to be refined to accommodate non-additive costs. However, due to multiplicative factors, it is not obvious if our cashing technique based on computing $U, D$ can be directly extended, especially if some of the coefficients are negative. In case of additive cost functions though, all of these schemes would correspond to our labeling scheme. Most of these structures pay the price in less efficient operators (such as apply operator) and larger memory requirements as they maintain more information. Therefore, for compiling Boolean functions, using these structures would pose an unnecessary overhead in comparison to OBDDs. Hence, for models with a large number of propositional (configuration) constraints, and an additive cost function, we would not gain from compiling using these structures even in the Boolean case. When the cost function is non-additive, introducing more elaborate cost representations might prove beneficial for reducing memory requirements, but might make our label computing technique unapplicable. From a practical point of view, while there are implementations supporting Boolean versions of these structures, we are not aware of any tool supporting multi-valued generalizations of these structures nor input language format that can be used for specifying general propositional constraints.

## 7.2 Multi-Objective Cost Processing

Our multiple-cost configuration is close to approaches within a framework of *multi-criteria optimization* where a decision maker should find a solution subject to multiple (often conflicting) objectives (Figueira et al., 2005; Ehrgott & Gandibleux, 2000). In particular, our MDD-based algorithms are very close to the approaches for solving *multiobjective shortest path problem*, where for a given graph $(V, E)$ each arc is labeled with multiple costs, and the goal is typically to compute the set of *Pareto-optimal* (efficient, non-dominated) solutions (Ehrgott & Gandibleux, 2000; Müller-Hannemann & Weihe, 2001; Tarapata, 2007; Reinhardt & Pisinger, 2009). It has been shown that the multi-objective shortest path problem is intractable. In particular, the number of Pareto-optimal solutions can grow exponentially with the number of vertices $|V|$, but a FPTAS (fully polynomial time approximation scheme) has been developed for approximating the set of Pareto-optimal solutions. However, the way in which the solution space of multi-criteria optimization problems is explored is significantly different from our approach. Typically, in each interaction step a subset of Pareto-optimal solutions is computed and afterwards a decision maker interactively navigates through the set in order to reach the satisfying compromising solution. Interactive methods in multi-criteria optimization usually compute a subset of solutions on the efficient frontier, suggest it to the user for evaluation, and based on his input compute a new set of solutions (Figueira et al., 2005, Chapter 16). These techniques would use the user input to better estimate the way to aggregate multiple objectives, and some of them would require the user to explicitly assign weights of importance to objectives. In contrast, instead of being primarily driven by the *costs* of solutions, our interactive approach supports reasoning about the variable assignments in the solutions themselves through valid domains computation. It is an inherently different way of exploring the solution space which is more adequate for users that want explicit control over variable assignments and not just indicating the importance of cost functions.

Most of the approaches in the CSP community model preferences as *soft constraints* (Meseguer, Rossi, & Shiex, 2006) that can be partially satisfied or violated, with a goal to find the most satisfying or the least violating solution. This usually presupposes that preferences can be aggregated via algebraic operators, and as such is more related to single-cost optimization problems. However, the approach by Rollón and Larrosa (2006) deals with multiple costs explicitly. It utilizes global structure (i.e. variable independencies) of the *weighted CSP* model to compute an efficient frontier through bucket-based variable elimination. A highly related approach that utilizes global structure of the *generalized additive independence* (GAI) network is presented by Dubus, Gonzales, and Perny (2009). In order to compute the efficient frontier, the authors use a message passing computation mechanism which is analogous to computing buckets. In addition, the authors develop a fully polynomial approximation scheme to approximate the efficient frontier and demonstrate the significant improvement in performance. However, neither of these methods can exploit the fact that the solution space of hard constraints is available in a compiled representation. Instead, these methods operate only over an unprocessed model specification (whether it is a weighted CSP or a GAI network) treating both the hard and soft constraints uniformly and hence allowing the scope of hard constraints to decrease the variable independencies in the model (and thus decrease the performance of the algorithms). Furthermore, the result of computation of these methods does not allow a full exploration of efficient solutions. For each value on the frontier only a single supporting efficient solution is maintained while we maintain for each efficient value the set of *all* supporting efficient solutions. Hence, it is not possible to efficiently retrieve valid domains even after the algorithms terminate. It would be interesting to see however, whether these methods could be adopted to work over MDD representations of a solution space.

Knapsack constraints are special case of two-cost configuration problems over a universally true MDD. Trick (2001) used dynamic programming to propagate Knapsack constraints during CSP search. Fahle and Sellmann (2002) presented an approximated filtering algorithm, based on various integer programming bounds for the Knapsack problem. Sellmann (2004) presented a fully polynomial time approximation algorithm for *approximated* filtering. However, these techniques were considered in constraint propagation context and none of them considered processing over existing MDD structure.

## 8. Conclusions and Future Work

In this paper we presented an extension of BDD-based interactive configuration to configuring in the presence of cost restrictions. We guarantee polynomial-time cost configuration when the cost function is *additive* and feasible solutions are represented using multi-valued decision diagram. We process cost restrictions over an MDD which is *extracted* from an underlying BDD. We therefore strictly extend BDD-based configuration of Hadzic et al. (2004) to support cost-bounding of additive cost functions without incurring exponential increase in complexity. Our implementation delivers running times that easily satisfy interactive response-time requirements. Furthermore, our approach can be extended to support bounding in the presence of *non-additive* and *semiring*-based costs.

We further extended our approach by considering cost bounding wrt. multiple costs. We proved that this is an NP-hard problem in the input MDD size even when processing only two linear inequalities with positive coefficients and Boolean variables. However, we provided a pseudo-polynomial scheme and fully polynomial approximation scheme for two-cost configuration (which, in principle, can be extended to any $k$-cost configuration for a fixed $k$). Our empirical evaluation demonstrated that despite inherent hardness of this problem we can still provide satisfying performance in interactive setting. Our interaction based on computing valid domains wrt. multiple cost restrictions is a novel addition to interaction modes within multiple-criteria decision making (Figueira et al., 2005). We provide an explicit control over variable assignments as well as cost functions.

In the future we plan to investigate other compiled representations over which delivering cost configuration might be efficient and to investigate practical approaches to processing non-unary cost functions. In particular, we plan to examine whether existing methods to multiobjective non-unary optimization (e.g., Rollón & Larrosa, 2006; Dubus et al., 2009) can be adopted to operate over MDD representation of a solution space.

## Acknowledgments

We would like to thank the anonymous reviewers for their extensive comments that helped us improve the paper. We would also like to thank Erik van der Meer for providing the T-shirt example. The first version of this paper was created while Tarik Hadzic was at the IT University of Copenhagen, while the updated version was made at the Cork Constraint Computation Centre with a support from an IRCSET/Embark Initiative Postdoctoral Fellowship Scheme.

## References

Amilhastre, J., Fargier, H., & Marquis, P. (2002). Consistency Restoration and Explanations in Dynamic CSPs-Application to Configuration. *Artificial Intelligence*, *135*(1-2), 199–234.

Bahar, R., Frohm, E., Gaona, C., Hachtel, E., Macii, A., Pardo, A., & Somenzi, F. (1993). Algebraic decision diagrams and their applications. In *IEEE/ACM International Conference on CAD*, pp. 188–191.

Bartzis, C., & Bultan, T. (2003). Construction of efficient BDDs for bounded arithmetic constraints. In Garavel, H., & Hatcliff, J. (Eds.), *TACAS*, Vol. 2619 of *Lecture Notes in Computer Science*, pp. 394–408. Springer.

Bodlaender, H. L. (1993). A tourist guide through treewidth. *Acta Cybernetica*, *11*, 1–23.

Bollig, B., & Wegener, I. (1996). Improving the variable ordering of OBDDs is NP-complete. *Computers, IEEE Transactions on*, *45*(9), 993–1002.

Bryant, R. E. (1986). Graph-Based Algorithms for Boolean Function Manipulation. *IEEE Transactions on Computers*, *35*, 677–691.

Bryant, R. E., & Chen, Y.-A. (1995). Verification of Arithmetic Circuits with Binary Moment Diagrams. In *In Proceedings of the 32nd ACM/IEEE Design Automation Conference*, pp. 535–541.

Darwiche, A., & Marquis, P. (2002). A Knowledge Compilation Map. *Journal of Artificial Intelligence Research*, *17*, 229–264.

Darwiche, A. (2002). A Logical Approach to Factoring Belief Networks. In Fensel, D., Giunchiglia, F., McGuinness, D., & Williams, M.-A. (Eds.), *KR2002: Principles of Knowledge Representation and Reasoning*, pp. 409–420 San Francisco, California. Morgan Kaufmann.

Darwiche, A. (2003). A differential approach to inference in Bayesian networks. *Journal of the ACM*, *50*(3), 280–305.

Dechter, R. (1999). Bucket Elimination: A Unifying Framework for Reasoning. *Artificial Intelligence, 113*(1-2), 41–85.

Dechter, R., & Pearl, J. (1989). Tree-Clustering for Constraint Networks. *Artificial Intelligence, 38*(3), 353–366.

Drechsler, R., Sarabi, A., Theobald, M., Becker, B., & Perkowski, M. A. (1994). Efficient representation and manipulation of switching functions based on ordered Kronecker functional decision diagrams. In *DAC '94: Proceedings of the 31st annual conference on Design automation*, pp. 415–419 New York, NY, USA. ACM.

Drechsler, R. (2001). Binary decision diagrams in theory and practice. *International Journal on Software Tools for Technology Transfer (STTT), 3*(2), 112–136.

Drechsler, R., Becker, B., & Ruppertz, S. (1996). K*BMDs: A New Data Structure for Verification. In *EDTC '96: Proceedings of the 1996 European conference on Design and Test*, p. 2 Washington, DC, USA. IEEE Computer Society.

Dubus, J.-P., Gonzales, C., & Perny, P. (2009). Multiobjective Optimization using GAI Models. In Boutilier, C. (Ed.), *IJCAI*, pp. 1902–1907.

Ehrgott, M., & Gandibleux, X. (2000). A Survey and Annotated Bibliography of Multiobjective Combinatorial Optimization. *OR Spektrum, 22*, 425–460.

Fahle, T., & Sellmann, M. (2002). Cost Based Filtering for the Constrained Knapsack Problem. *Annals of Operations Research, 115*, 73–93.

Fargier, H., & Marquis, P. (2006). On the Use of Partially Ordered Decision Graphs in Knowledge Compilation and Quantified Boolean Formulae. In *Proceedings of AAAI 2006*, pp. 42–47.

Fargier, H., & Marquis, P. (2007). On Valued Negation Normal Form Formulas. In *Proceedings of IJCAI 2007*, pp. 360–365.

Fargier, H., & Vilarem, M.-C. (2004). Compiling CSPs into Tree-Driven Automata for Interactive Solving. *Constraints, 9*(4), 263–287.

Figueira, J. R., Greco, S., & Ehrgott, M. (2005). *Multiple Criteria Decision Analysis: State of the Art Surveys*. Springer Verlag, Boston, Dordrecht, London.

Garey, M. R., & Johnson, D. S. (1979). *Computers and Intractability-A Guide to the Theory of NP-Completeness*. W H Freeman & Co.

Hadzic, T., Subbarayan, S., Jensen, R. M., Andersen, H. R., Møller, J., & Hulgaard, H. (2004). Fast Backtrack-Free Product Configuration using a Precompiled Solution Space Representation. In *In Proceedings of PETO Conference*, pp. 131–138. DTU-tryk.

Hadzic, T., & Andersen, H. R. (2006). A BDD-based Polytime Algorithm for Cost-Bounded Interactive Configuration. In *Proceedings of AAAI 2006*, pp. 62–67.

Hadzic, T., Hansen, E. R., & O'Sullivan, B. (2008). On Automata, MDDs and BDDs in Constraint Satisfaction. In *Proceedings of the ECAI 2008 Workshop on Inference Methods based on Graphical Structures of Knowledge*.

Huang, J., & Darwiche, A. (2004). Using DPLL for efficient OBDD construction. In *Proceedings of SAT 2004*, pp. 127–136.

Huang, J., & Darwiche, A. (2005). DPLL with a trace: From SAT to knowledge compilation. In Kaelbling, L. P., & Saffiotti, A. (Eds.), *IJCAI*, pp. 156–162. Professional Book Center.

Ibarra, O., & Kim, C. (1975). Fast approximation algorithms for the knapsack and sum of subset problem. *Journal of the ACM, 22*, 463–468.

Jensen, R. M. (2007). CLab: A C++ library for fast backtrack-free interactive product configuration. `http://www.itu.dk/people/rmj/clab/`.

Kebschull, U., & Rosenstiel, W. (1993). Efficient graph-based computation and manipulation of functional decision diagrams. *Design Automation, 1993, with the European Event in ASIC Design. Proceedings. [4th] European Conference on*, 278–282.

Kellerer, H., Pferschy, U., & Pisinger, D. (2004). *Knapsack Problems*. Springer, Berlin, Germany.

Lai, Y.-T., & Sastry, S. (1992). Edge-valued binary decision diagrams for multi-level hierarchical verification. In *DAC '92: Proceedings of the 29th ACM/IEEE conference on Design automation*, pp. 608–613 Los Alamitos, CA, USA. IEEE Computer Society Press.

Lichtenberg, J., Andersen, H. R., Hulgaard, H., Møller, J., & Rasmussen, A. S. (2001). Method of configuring a product. US Patent No: 7,584,079.

Lind-Nielsen, J. (2001). BuDDy - A Binary Decision Diagram Package. `http://sourceforge.net/projects/buddy`.

Mateescu, R., Dechter, R., & Marinescu, R. (2008). AND/OR Multi-Valued Decision Diagrams (AOMDDs) for Graphical Models. *Journal of Artificial Intelligence Research, 33*, 465–519.

Meinel, C., & Theobald, T. (1998). *Algorithms and Data Structures in VLSI Design*. Springer.

Meseguer, P., Rossi, F., & Shiex, T. (2006). Soft constraints. In Rossi, F., van Beek, P., & Walsh, T. (Eds.), *Handbook of Constraint Programming*, Foundations of Artificial Intelligence, chap. 9, pp. 281–328. Elsevier Science Publishers, Amsterdam, The Netherlands.

Miller, D. M., & Drechsler, R. (2002). On the Construction of Multiple-Valued Decision Diagrams. In *Proceedings of the 32nd International Symposium on Multiple-Valued Logic (ISMVL'02)*, p. 245 Washington, DC, USA. IEEE Computer Society.

Møller, J., Andersen, H. R., & Hulgaard, H. (2002). Product configuration over the internet. In *INFORMS Conference on Information Systems and Technology*.

Müller-Hannemann, M., & Weihe, K. (2001). Pareto Shortest Paths is Often Feasible in Practice. In *WAE '01: Proceedings of the 5th International Workshop on Algorithm Engineering*, pp. 185–198 London, UK. Springer-Verlag.

Nicholson, R., Bridge, D. G., & Wilson, N. (2006). Decision Diagrams: Fast and Flexible Support for Case Retrieval and Recommendation. In *Proceedings of ECCBR 2006*, pp. 136–150.

Pargamin, B. (2003). Extending Cluster Tree Compilation with non-Boolean variables in Product Configuration: a Tractable Approach to Preference-based Configuration. In *IJCAI'03 Workshop on Configuration*.

Reinhardt, L. B., & Pisinger, D. (2009). Multi-Objective and Multi-Constrained Non-Additive Shortest Path Problems. Computers and Operations Research. Submitted. Technical report version available at: `http://man.dtu.dk/upload/institutter/ipl/publ/publikationer%202009/rapport%2016.pdf`.

Rollón, E., & Larrosa, J. (2006). Bucket elimination for multiobjective optimization problems. *Journal of Heuristics*, *12*(4-5), 307–328.

Sanner, S., & McAllester, D. A. (2005). Affine Algebraic Decision Diagrams (AADDs) and their Application to Structured Probabilistic Inference. In *Proceedings of IJCAI 2005*, pp. 1384–1390.

Schuurman, P., & Woeginger, G. J. (2005). Approximation schemes — a tutorial. In Moehring, R., Potts, C., Schulz, A., Woeginger, G., & Wolsey, L. (Eds.), *Lectures on Scheduling*. Forthcoming.

Sellmann, M. (2004). The Practice of Approximated Consistency for Knapsack Constraints. In McGuinness, D. L., & Ferguson, G. (Eds.), *AAAI*, pp. 179–184. AAAI Press / The MIT Press.

Somenzi, F. (1996). CUDD: Colorado university decision diagram package. `ftp://vlsi.colorado.edu/pub/`.

Srinivasan, A., Kam, T., Malik, S., & Brayton, R. K. (1990). Algorithms for discrete function manipulation. In *International Conference on CAD*, pp. 92–95.

Subbarayan, S., Jensen, R. M., Hadzic, T., Andersen, H. R., Hulgaard, H., & Møller, J. (2004). Comparing two implementations of a complete and backtrack-free interactive configurator. In *Proceedings of CP'04 CSPIA Workshop*, pp. 97–111.

Subbarayan, S. M. (2008). *On Exploiting Structures for Constraint Solving*. Ph.D. thesis, IT University of Copenhagen, Copenhagen.

Tafertshofer, P., & Pedram, M. (1997). Factored edge-valued binary decision diagrams. In *Formal Methods in System Design*, Vol. 10, pp. 243–270. Kluwer.

Tarapata, Z. (2007). Selected multicriteria shortest path problems: An analysis of complexity, models and adaptation of standard algorithms. *International Journal of Applied Mathematics and Computer Science, 17*(2), 269–287.

Trick, M. (2001). A dynamic programming approach for consistency and propagation for knapsack constraints. In *3rd international workshop on integration of AI and OR techniques in constraint programming for combinatorial optimization problems CP-AI-OR*, pp. 113–124.

Vempaty, N. R. (1992). Solving constraint satisfaction problems using finite state automata. In *Proceedings of the Tenth National Conference on Artificial Intelligence*, pp. 453–458.

Walsh, T. (2000). SAT v CSP. In Dechter, R. (Ed.), *Proceedings of CP 2000*, Lecture Notes in Computer Science, pp. 441–456.

Wegener, I. (2000). *Branching Programs and Binary Decision Diagrams*. Society for Industrial and Applied Mathematics (SIAM).

Wilson, N. (2005). Decision diagrams for the computation of semiring valuations. In *Proceedings of the Nineteenth International Joint Conference on Artificial Intelligence (IJCAI-05)*, pp. 331–336.